%% file: main.tex
\documentclass[10pt,twocolumn,letterpaper]{article}

\usepackage[pagenumbers]{iccv} % To force page numbers, e.g. for an arXiv version


\definecolor{iccvblue}{rgb}{0.21,0.49,0.74}
\usepackage[pagebackref,breaklinks,colorlinks,allcolors=iccvblue]{hyperref}

\usepackage{adjustbox}
\usepackage{multirow}
\usepackage{booktabs}

\usepackage{xcolor}         % colors
\definecolor{darkgreen}{rgb}{0.0, 0.75, 0.0} % Dark green
\definecolor{darkred}{rgb}{0.8, 0.0, 0.0}   % Dark red

\usepackage{pifont}   % Provides \ding symbols

\newcommand{\cmark}{\textcolor{darkgreen}{\ding{51}}} % Green check mark
\newcommand{\xmark}{\textcolor{darkred}{\ding{55}}}   % Red cross mark

\def\paperID{14376} % *** Enter the Paper ID here
\def\confName{ICCV}
\def\confYear{2025}

\title{Att-Adapter: A Robust and Precise  Domain-Specific Multi-Attributes T2I Diffusion Adapter via Conditional Variational Autoencoder}

\author{\normalsize Wonwoong Cho\textsuperscript{1}\thanks{A significant portion of this work was performed during internship at Toyota Research Institute.}
\and
\normalsize Yan-Ying Chen\textsuperscript{2}
\and
\normalsize Matthew Klenk\textsuperscript{2}
\and
\normalsize David I. Inouye\textsuperscript{1}
\and
\normalsize Yanxia Zhang\textsuperscript{2}
\and
\normalsize \textsuperscript{1} Purdue University, West Lafayette, IN, 47906
\and
\normalsize \textsuperscript{2} Toyota Research Institute, Los Altos, CA, 94022
}
\begin{document}

% \institute{Purdue University, West Lafayette IN 47907, USA \and
% Toyota Research Institute, Los Altos CA 95110, USA
% }

\maketitle
\input{sec/0_abstract}    
\input{sec/1_intro}

\input{sec/2_relatedwork}
\input{sec/3_method}

\input{sec/4_experiments}

\input{sec/5_conclusion}

\clearpage
\section*{Acknowledgement}
We would like to express our gratitude to Kalani Murakami for his assistance and support for processing EVOX data during the course of this research.

D.I. acknowledges support from ARL (W911NF-2020-221). Any opinions, findings, and conclusions or recommendations expressed in this material are those of the authors and do not necessarily reflect the views of the sponsor. 

{
    \small
    \bibliographystyle{ieeenat_fullname}
    \bibliography{main}
}

\clearpage

\input{rebuttal}

\end{document}

%% file: sec/0_abstract.tex
\begin{abstract}
Text-to-Image (T2I) Diffusion Models have achieved remarkable performance in generating high quality images. However, enabling precise control of continuous attributes, especially multiple attributes simultaneously, in a new domain (e.g., numeric values like eye openness or car width) with text-only guidance remains a significant challenge. 
To address this, we introduce the \textbf{Attribute (Att) Adapter}, a novel plug-and-play module designed to enable fine-grained, multi-attributes control in pretrained diffusion models. Our approach learns a single control adapter from a set of sample images that can be unpaired and contain multiple visual attributes. The Att-Adapter leverages the decoupled cross attention module to naturally harmonize the multiple domain attributes with text conditioning.
We further introduce Conditional Variational Autoencoder (CVAE) to the Att-Adapter to mitigate overfitting, matching the diverse nature of the visual world.
Evaluations on two public datasets show that Att-Adapter outperforms all LoRA-based  baselines in controlling continuous attributes. 
Additionally, our method enables a broader control range and also improves disentanglement across multiple attributes, surpassing StyleGAN-based techniques. Notably, Att-Adapter is flexible, requiring no paired synthetic data for training, and is easily scalable to multiple attributes within a single model. The project page is available at: \url{https://tri-mac.github.io/att-adapter/}.
\end{abstract}

%% file: sec/1_intro.tex
\newcommand{\david}[1]{{\color{red}[#1]}}

\section{Introduction}
\label{sec:intro}
\begin{figure}[t]
  % \vspace{-0.2in}
  \centering
    \includegraphics[height=.45\linewidth]{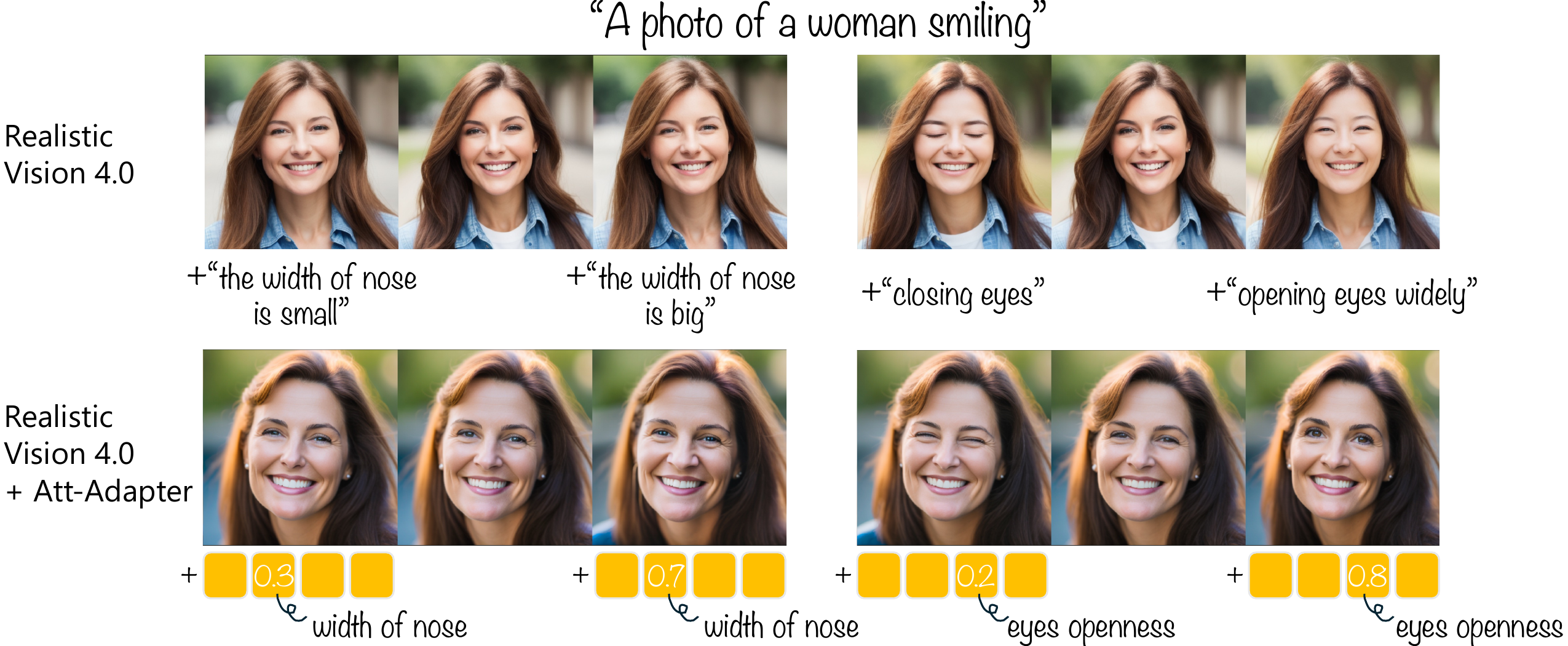}
  \caption{Att-Adapter enables continuous control (e.g., 0 to 1) over subtle visual details like nose width or eye openness, which are difficult to capture with discrete text-based conditioning.
  }
  \label{fig:motivating example 1}
  \vspace{-0.2in}
\end{figure}

Large pretrained T2I diffusion models
%like CLIP~\cite{radford2021learning}, Text-to-Image (T2I) diffusion models
~\cite{rombach2021highresolution,ramesh2021zero,nichol2021glide,saharia2022photorealistic} have achieved remarkable success in generating high quality and realistic images based solely on text prompts. 
%\david{Control/adapt to a specific domain instead of general conditional generation.} 
However, allowing users precise control over the generated content or tailoring it to a specific new domain remains challenging \cite{gal2022textual,gu2024mix,ruiz2022dreambooth,yang2024lora}. For example, with the prompt ``A photo of a woman smiling'', one might want to adjust specific facial attributes such as ``width of nose" and ``openness of eyes" by assigning specific numeric values to these attributes. As Fig.~\ref{fig:motivating example 1} demonstrates, current models like Stable Diffusion~\cite{rombach2021highresolution} (Realistic Vision 4.0\footnote{\url{https://huggingface.co/stablediffusionapi/realistic-vision-v40}}) fail to reliably control these fine-grained attributes, often misinterpreting or neglecting the specified modifications. 

Early domain adaptation methods rely on text-based conditioning, either by optimizing text tokens \cite{ruiz2022dreambooth, gal2022textual, zhang2023iti} or fine-tuning T2I models using Low-Rank Adaptation (LoRA) \cite{hu2022lora}. However, text-only conditioning are inherently limited: texts often underspecifies visual details such as object type, perspective and style \cite{hutchinson2022underspecification, zhang2023iti}, and existing methods typically handle only discrete attribute values. 

\begin{figure}[t]
  % \vspace{-0.2in}
  \centering
    \includegraphics[height=2in]{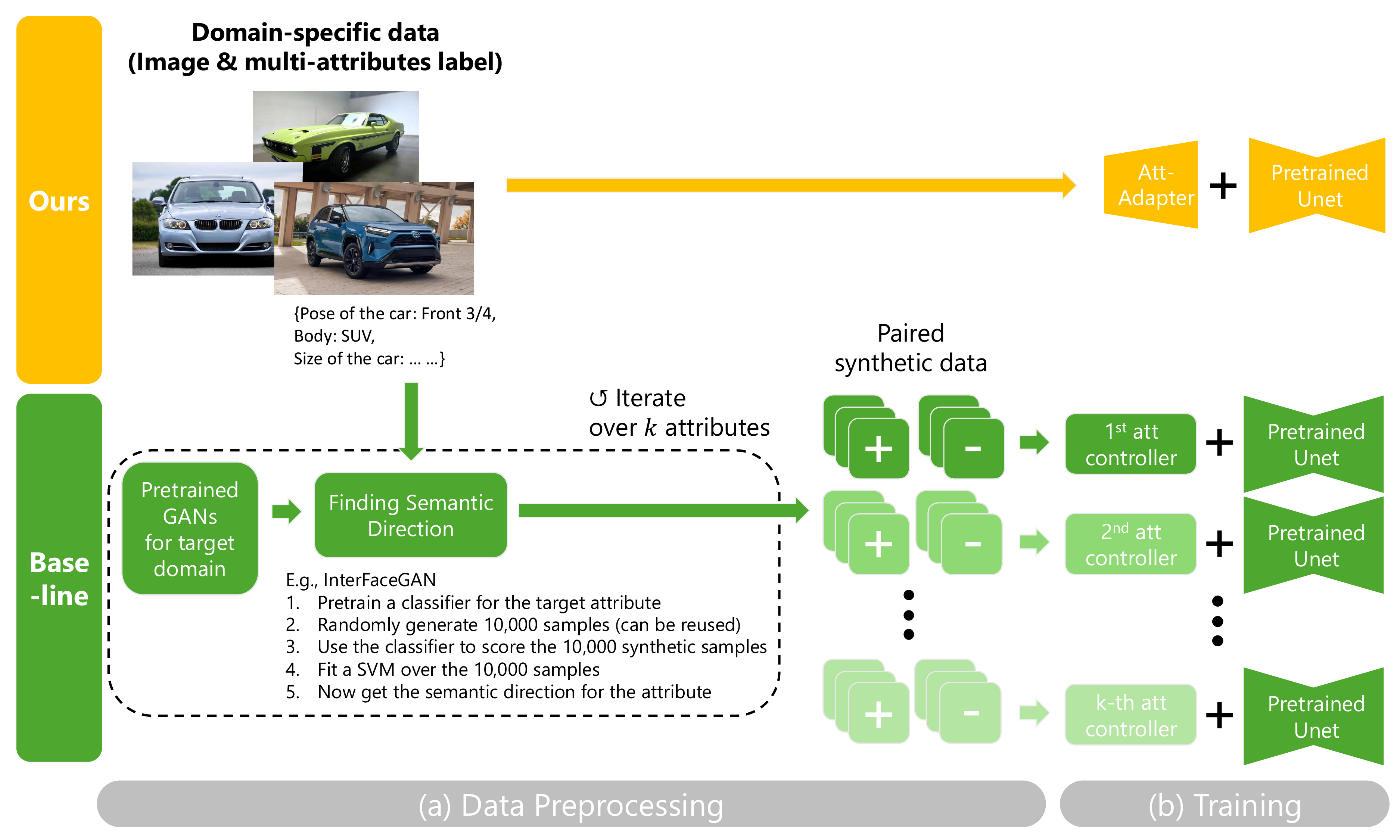}
  \caption{\textbf{Workflow comparison.} Unlike ConceptSlider~\cite{gandikota2024concept}, which demands paired data and separate models for each attribute, our approach supports unpaired multi-attribute control in a single framework. See Sec.~\ref{sec:related_works} for details.
  %Performance comparisons between the pretrained diffusion models and our proposed method in controlling domain specific knowledge.
  }
  \label{fig:motivating example 2}
  \vspace{-0.1in}
\end{figure}

Recent efforts attempt to address continuos attribute \cite{baumann2025attributecontrol, li2024stylegan, parihar2024precisecontrol, gandikota2024concept}. AttributeControl \cite{baumann2025attributecontrol} modifies CLIP embeddings but struggles with indescribable visual attributes (e.g., nose width) that rarely appear in text. Other methods, such as $W_+$ Adapter\cite{li2024stylegan} and PreciseControl \cite{parihar2024precisecontrol}, depend on pretrained StyleGAN latent spaces, making generation quality highly dependent on separately trained GANs. ConceptSlider \cite{gandikota2024concept} proposes a LoRA-based controller for indescribable visual attributes but requires paired data and trains separate models per attribute (Fig.~\ref{fig:motivating example 2}). This limits its use to the case where only one attribute is paired while others remain unchanged, which makes it infeasible in unpaired multi-attribute data settings without generating synthetic pairs. However, generating synthetic paired data adds extra preprocessing steps and restricts performance to the quality of the data generator (e.g., StyleGAN). A detailed comparison with existing methods is listed in Table \ref{table: comparing the settings with baselines}.

% Despite recent advances in domain-specific attribute controls over pretrained diffusion models, there are two remaining challenges. First, obtaining domain-specific $\it{paired}$ data is challenging. For instance, acquiring two Lamborghini cars with different body types while keeping all other features identical is difficult. % Similarly, obtaining two face images with same ID but different nose shape is inherently impossible.  

\begin{table}[t]
\centering
\begin{adjustbox}{width=.48\textwidth}
\begin{tabular}{l|c|cccc}
\multicolumn{1}{c|}{} & \begin{tabular}[c]{@{}c@{}}Pretrained \\ GANs\end{tabular} & \begin{tabular}[c]{@{}c@{}}Continuous\\ Attrs\end{tabular} & \begin{tabular}[c]{@{}c@{}}Indescribable\\ Visual Attrs\end{tabular} & \begin{tabular}[c]{@{}c@{}}Multi-Attrs\\ Training\end{tabular} & \begin{tabular}[c]{@{}c@{}}Real Data\\ Training\end{tabular} \\ \hline
AttrCtrl~\cite{baumann2025attributecontrol}                                                              & \xmark                                                     & \cmark                                                     & \xmark                                                               & \xmark                      & \xmark                                   \\
$W_+$Adapter~\cite{li2024stylegan}                                                         & \cmark                                                     & \cmark                                                     & \cmark                                                               & \xmark                      & \xmark                                   \\
ConceptSlider~\cite{gandikota2024concept}                                                       & \cmark                                                     & \cmark                                                     & \cmark                                                               & \xmark                    & \xmark                                     \\
ITI-GEN~\cite{zhang2023iti}                                                                 & \xmark                                                     & \xmark                                                     & \cmark                                                               & \xmark           & \cmark                                               \\
LoRA~\cite{hu2022lora}                                                                 & \xmark                                                     & \xmark                                                     & \cmark                                                               & \cmark           & \cmark                                               \\
Ours                                                                 & \xmark                                                     & \cmark                                                     & \cmark                                                               & \cmark           & \cmark                                              
\end{tabular}

\end{adjustbox}
\caption{\textbf{Requirement and capability comparisons}. The leftmost column shows a requirement while the others show benefits. Multi-Attrs indicate many attributes such as 20. Most baselines rely on a pretrained latent concept space where the synthetic attribute values are not directly from the real-world data.}
\label{table: comparing the settings with baselines}
\vspace{-0.2in}
\end{table}

In this paper, we propose \textbf{Att-Adapter}, a novel plug-and-play adapter that enables precise control over multiple attributes in pretrained diffusion models. As illustrated in the top row of Fig.~\ref{fig:motivating example 2}, Att-Adapter learns multi-attribute control from unpaired real data by leveraging a decoupled cross-attention module~\cite{ye2023ip-adapter} to effectively fuse conditions from different modalities (e.g., view angle, size, style). We found that directly mapping multiple attributes with a naive MLP leads to overfitting (Fig.~\ref{fig:random sampling showing overfitting} and Table~\ref{table:ablation study without cvae}); To this end, we incorporate a Conditional Variational Autoencoder (CVAE)~\cite{sohn2015learning,zhao2017learning} to improve robustness against overfitting. By performing an approximate Bayesian inference~\cite{kingma2013auto}, the CVAE can naturally regularize Att-Adapter via its KL divergence term. This not only mitigates overfitting but also smooths the conditional representation, modeling the natural randomness for given attribute condition. % 

Extensive experiments on face and car datasets demonstrate that Att-Adapter outperforms state-of-the-art methods in both continuous and discrete attribute control. Our contributions are as follows:
\begin{itemize}[leftmargin=1em]
  \item \textbf{Unpaired Multi-Attribute Control:} Att-Adapter simultaneously controls multiple attributes from unpaired data, eliminating the need for costly paired samples.
  \item \textbf{Continuous and Indescribable Attribute Manipulation:} Our approach precisely controls subtle visual details that are difficult to describe textually.
  \item \textbf{Robust Control:} The integration of a CVAE mitigates overfitting, ensuring stable and accurate attribute control.
\end{itemize}

%Furthermore, our results demonstrate that our approach can maintain pretrained knowledge (i.e., avoid catastrophic forgetting) so that the free-form text prompt and conditioning can be combined seamlessly.

%Additionally, we conduct experiments demonstrating that our model can balance pretrained knowledge with domain-specific knowledge during image generation. Using a user-defined parameter, the model generates diverse images guided by text prompts, effectively addressing the challenge of catastrophic forgetting. Finally, we show that our approach can control attributes in a PCA feature space, even when these attributes are not explicitly defined in the design dataset. This capability allows our model to support various applications, such as design, where attributes (e.g., style) exist in a lower-dimensional space compared to pixels. 

%% file: sec/2_relatedwork.tex
\section{Related Works}
\label{sec:related_works}
Additional related works on controllable T2I models, such as IP-Adapter~\citep{ye2023ip-adapter} are in Section~\ref{Appendix_sec:additional related works} in the Appendix.

\vspace{0.1in}
\noindent\textbf{Single Attribute Control for T2I Generation.}
Diffusion models (DMs) \cite{dhariwal2021diffusion, ho2020denoising, sohl2015deep} are one of the widely-used text-to-image generative models due to its impressive performance.
To customize DMs for a particular domain or new concept, fine-tuning the U-Net is a common approach \cite{ruiz2022dreambooth, kumari2023multi}, although it can lead to catastrophic forgetting~\cite{mccloskey1989catastrophic,luo2023empirical}. Instead of retraining all model parameters, Low-Rank Adaptations (LoRA) \cite{hu2022lora}, adds trainable low rank decomposition matrices to each layer of a frozen model, significantly reducing trainable parameters while maintaining performance. Other methods, such as  DreamBooth \cite{ruiz2022dreambooth} and Textual Inversion \cite{gal2022textual}, modify text embeddings of pretrained T2I models \cite{ruiz2022dreambooth, gal2022textual, zhang2023iti}. %For example, DreamBooth \cite{ruiz2022dreambooth} fine-tunes DMs with a few images of the subject and text prompts containing a unique identifier and category of the subject, which enables generating subjects in diverse unseen contexts. Textual Inversion \cite{gal2022textual} associates new concept with special tokens in the text prompt, enabling concept generation similar to the given examples. 
%LoRA can combine with text embedding methods for efficiency, but 
Both LoRA and text embedding approaches require training a new LoRA network or token for each attribute. Additionally, they struggle to represent continuous attribute values without discretization.

\vspace{0.1in}
\noindent\textbf{Multi-Attribute Control for T2I Generation.}
Existing methods often combine individually trained LoRAs through weighted averaging, which can lead to concept vanish or distortion \cite{luo2023empirical, kumari2023multi, yang2024lora, gu2024mix}. Approaches like Mix-of-Show \cite{gu2024mix} and Lora-composer \cite{yang2024lora} attempt to address this but still require separate LoRAs, limiting scalability and forcing discretization of continuous attributes.

Text embedding modification methods face similar challenges. Custom Diffusion \cite{kumari2023multi} uses unique tokens for each concept, optimizing multiple embeddings but lacking continuous control and extrapolation. DreamDistribution \cite{zhao2023dreamdistribution} learns prompts from reference images, creating diverse outputs but without fine control over continuous attributes. ITI-GEN \cite{zhang2023iti} addresses catastrophic forgetting with prompt engineering, using pretrained parameters for consistency and balance sampling. These methods struggle in representing continuous attribute and require discretization. Furthermore, the token embeddings need to be trained with all the attribute combinations, significantly limiting its scalability. 
%Their training process exceeds 48 GB VRAM (RTX 6000) with combinations of 3 attributes with 10 categories (i.e., $10^3$). Our main experiments deal with 20 attributes with 10 categories $10^{20}$. %ITI-GEN also cannot take text prompt during the inference, which limits their capacity as an adapter for pretraind diffusion models.

\vspace{0.1in}
\noindent\textbf{Continuous Attribute Control.}
Several methods aim to enable attribute control on a continuous scale \cite{baumann2025attributecontrol, li2024stylegan, parihar2024precisecontrol, gandikota2024concept}. Pretraind Generative Adversarial Networks (GANs)~\cite{goodfellow2020generative} such as StyleGAN~\cite{karras2019style} have been extensively explored for continuous attribute control. For example,  InterFaceGAN~\cite{shen2020interfacegan}, GANSpace~\cite{harkonen2020ganspace}, and StyleSpace~\cite{wu2021stylespace} can be used to enable the attribute controls over the pretrained GAN's space. %We used InterFaceGAN in our experiments due to the interpretability of the semantic changes and the attribute retrieval simplicity; we find the target attribute instead of using the pre-computed because of the limited range of options.
$W_+$ Adapter~\cite{li2024stylegan} and PreciseControl~\cite{parihar2024precisecontrol} link pretrained diffusion models with the pretrained GANs' $W_+$ space~\cite{shen2020interfacegan,harkonen2020ganspace,wu2021stylespace}. However, their performance is constrained by the quality of the $W_+$ space. Furthermore, if there is no public pretrained GANs for specific domains, they require training new GANs from scratch. 

ConceptSlider~\cite{gandikota2024concept} trains LoRA adapters for fine-grained control of indescribable visual concepts. Unlike ConceptSlider, our method supports multi-attribute control in both training and sampling without requiring separate models for each attribute.
Moreover, ConceptSlider relies on paired data, which is often synthesized and can degrade performance when used in unpaired settings. In contrast, our approach remains effective with real data, making it more flexible and practical. AttributeControl~\cite{baumann2025attributecontrol} proposes learning a certain attribute (e.g., old) in the CLIP text embedding space by contrasting the attribute-related words (e.g., old and young). However, their method is inherently not suitable for attributes that are difficult to describe in text. In contrast to existing works, our method does not rely on paired data and a pretrained latent embedding space. 

%% file: sec/3_method.tex
\section{Method}
\label{sec:method}
% why not? Leverage pretrained feature space in IP adapter. more general for both discrete and continuous variables. Learn a lower dimensional representation. Key thing is to learn mapping is from V->C. 
% Given (X,C) pairs
% Convert this to (V,C) pairs via pretrained IP adapter.
% Learn p(V|C)
% V|C ~ N(mu(C), sigma(C))
% \david{Very rough outline brainstorm idea of motivation storyline (focus on continuous attributes):
% \begin{enumerate}
%     \item Because we focus on control of continuous attributes, we cannot directly modify the text conditioning module as in prior methods.
%     \item We will need another way to inject conditioning knowledge (also domain-specific knowledge?).
%     \item We choose to use a setup similar to IP-Adapter that puts a parallel conditioning path to text conditioning.
%     \item Because directly conditioning is unlikely to work because ....?, we decide to use the representation space of a pretrained IP-adapter as a starting point.
%     \item Additionally, because we also want to condition on discrete attributes, it is natural to have a distribution over conditioning values for any given conditioning vector $C$, so we consider a probabilistic conditional representation.
%     \item This leads us to use a conditional VAE using pretrained IP-adapter modules where the focus is learning the conditional generative model but using the pretrained IP-adapter space to find a relatively good representation space.
% \end{enumerate}
% }

% \subsection{Preliminary Knowledge}
% Diffusion Models, IP-Adapter, IP-Adapter FaceID
In this section, we first describe the naive version of Att-Adapter with a simple MLP network, where personal identity overfitting is observed. We then describe Att-Adapter with conditional variational autoencoder (CVAE) to mitigate the undesirable overfitting issue.
\subsection{Problem Formulation}

Att-Adapter takes a set of domain-specific attributes $C$ as input with the text prompt $Y$, modeling $p(X|Y, C)$. Our method is designed to support multiple attributes. Hence, we can scale to an unlimited number of conditions with negligible increase in memory usage. For example, $C_1$, $C_2$, $C_3$ and $C_4$ could be `height of eyes', `width of nose', `age', and `probability of Asian', and all of the attributes can be trained altogether, which can be represented as $p(X|Y,C_1,C_2,C_3,C_4)$.

Formally, similar to previous works~\cite{ho2022classifier,song2020score}, we can approximate $p(X|Y, C)$ with the reverse diffusion process with timestep $t$ by:
\begin{align}
 \nabla_{X} &\log p(X(t)|Y,C) \approx \\
 &\epsilon_\theta(X(t)) + w \left( \epsilon_\theta(X(t),Y,C) - \epsilon_\theta(X(t)) \right) \nonumber
\end{align}
where $\epsilon_\theta(X(t))$ and $\epsilon_\theta(X(t),Y,C)$ are unconditional and conditional diffusion models, respectively. Now, the question boils down to how to implement $\epsilon_\theta(X(t),Y,C)$.

\subsection{Att-Adapter}
\label{subsec:att-adapter}

To model the conditional with diffusion models, i.e., $\epsilon_\theta(X(t),Y,C)$, we adopt the decoupled cross-attention module~\cite{ye2023ip-adapter} as it is highly effective in combining two different conditions with different modalities. %In order for training stability, we start from the pretrained parameters of IP-Adapter. 
Specifically, the decoupled cross-attention modules for attribute feature injection are formulated as:
%IP-Adapter is composed of a pretrained vision encoder (e.g., CLIP~\cite{radford2021learning} or a face recognition model~\cite{insightface,deng2018arcface,deng2019retinaface}), an image feature projection layer, and feature injection modules per denosing UNet layer. 

\begin{equation}
\text{Attention}(\textbf{\textmd{Q}},\textbf{\textmd{K}}_{\text{text}},\textbf{\textmd{V}}_{\text{text}})+ \lambda \text{Attention}(\textbf{\textmd{Q}},\textbf{\textmd{K}}_{\text{attr}},\textbf{\textmd{V}}_{\text{attr}}),    
\label{eq:decoupled cross attention}
\end{equation}
where $\lambda$ controls the strength of the attribute, and the key and value for attributes come from the multi-attributes condition through Att-Adapter, i.e., $\textbf{\textmd{K}}_{\text{attr}}=W_k \text{AttAdapter}(c)$ and $\textbf{\textmd{V}}_{\text{attr}}=W_v \text{AttAdapter}(c)$.

In the following sections, we first describe our Att-Adapter using a naive MLP-based mapping network and summarize our key observations. We then introduce our enhanced Att-Adapter with a CVAE module to overcome overfitting in multi-attribute control generation.  

\subsubsection{Naive Att-Adapter with a MLP Mapping}
%Fig.~\ref{fig:AttAdapter overview} (a) shows 
A naive version of Att-Adapter for multi-attributes control is to have a simple linear function $g$ that maps the multiple attributes $C$ into the input space of the pretrained decoupled cross-attention modules. More precisely, $g$ outputs the input of the layer norm~\cite{ba2016layer}, and the output of the norm layer is fed into the decoupled cross-attention module. The linear function $g$ and the decoupled cross-attention modules are optimized with the denoising objective function, i.e., \\
\begin{equation}
\mathcal{L}_{\text{denoising}}(x_t,t,y,c;g)=\mathbb{E}\left[ \| \epsilon - \epsilon_\theta(x_t,t,y,g(c)) \|^2 \right],   
\end{equation}
where $\epsilon_\theta$ is pretrained diffusion models~\cite{rombach2021highresolution}.

We find that the naive Att-Adapter can control multiple attributes and can be integrated well with text conditioning, benefiting from the decoupled cross-attention module to merge multimodal conditions. However, we observed overfitting when fixing attribute values $C=c$ while varying the random Gaussian sample at the timestep $T$. In this case, diffusion models tend to generate visually similar images (see w.o. CVAE inFig.~\ref{fig:random sampling showing overfitting} and Table~\ref{table:ablation study without cvae}). This is problematic since a single set of $c$ does not correspond to a unique human subject. Many individuals may share similar attributes such as nose width. To address this, we propose an enhanced Att-Adapter that overcomes overfitting while maintaining precise attribute control.

% To model $\epsilon_\theta(X(t),Y,C)$ effectively, one idea is to use pretrained IP-adapter~\cite{ye2023ip-adapter} to provide a separate image branch from the text branch, i.e., $\epsilon_\theta(X(t),Y,X')$, where $X'$ is an image prompt (see Fig.~\ref{fig:graphical visualizations of ipadapter and attadapter} (a)). Our motivation is that the image prompt embedding contains rich visual information which adds the indescribable details that effectively augment the text-only guidance. 

%In practice, however, having pretrained diffusion models taking both text and domain-specific attributes as input is not trivial because many of these attributes require visual representation to characterize fine-grained information. % as attributes information is non spatial\footnote{Most of the related works are designed for spatial conditioning such as depth map and segmentation mask, to the best of our knowledge.}.

% In order to replace $\epsilon_\theta(X(t),Y,X')$ to $\epsilon_\theta(X(t),Y,C)$, 
% In order to leverage the strong representation power of IP-Adapter, we inject $C$ to the intermediate feature representation of IP-Adapter. Specifically, we model a conditional distribution over $C$ because we need to condition on both continuous and discrete attributes.

\begin{figure*}[t]
    % \vspace{-0.3in}
     \centering
     \includegraphics[width=.9\linewidth]{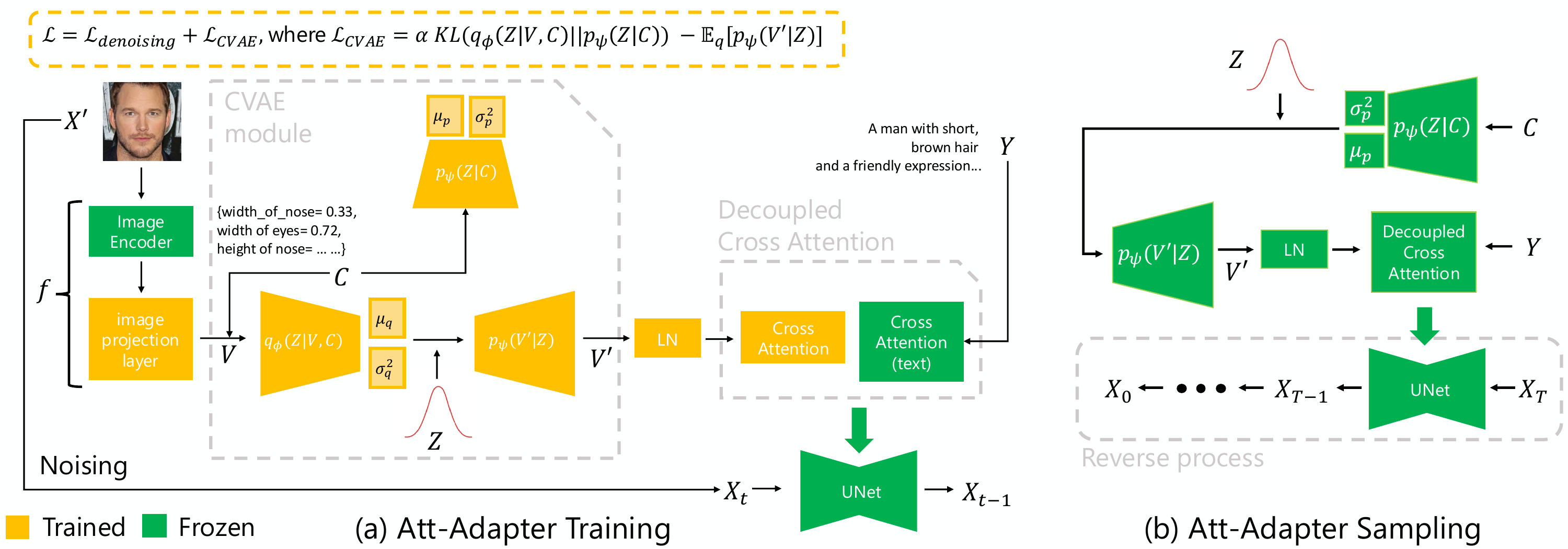}
     \caption{An illustration of the training process of Att-Adapter. (a) shows the training process for the Att-Adapter with CVAE. Our model learns a single multi-attribute control model given a set of sample images with different values for attribute $C$ (i.e., width of nose/eyes). (b) shows the sampling process of Att-Adapter. Our CVAE decoder $p_\psi(V'|Z)$ takes a random sample by conducting reparameterization trick from the conditional prior $p_\psi(Z|C)$ taking $C$ as input and having been learned to output the mean and the variance of the isotropic Gaussian. We empirically show that our CVAE modeling is effective in preventing overfitting as we intended.} % Both (a) and (b) can augment the pretrained diffusion models with the conditional domain-specific multi-attributes $C$. However, the simple setting (a) suffers from overfitting while (b) does not have such issue by introducing CVAE objective regularizing the training process against overfitting and sampling from the conditional prior distribution.}
     \label{fig:AttAdapter overview}
     \vspace{-0.1in}
\end{figure*}

\subsubsection{Att-Adapter with CVAE}
To mitigate the undesirable overfitting, we design Att-Adapter as a one-to-many mapping function from $C$. To this end, we introduce Conditional Variational Autoencoder~\cite{sohn2015learning,zhao2017learning} (CVAE) that can naturally regularize the model parameters from overfitting and can learn the latent space $Z$. To be specific, the KL-divergence term with the prior distribution can make the conditional latent space smoother, preventing the over-simplified representation for $C$. This is because VAEs~\cite{kingma2013auto} perform approximate Bayesian inference, promoting generalization rather than memorizing or simplifying the data distribution. %In other words, during the sampling, we do not need to forward an image prompt but can give only a few dimensional vector (i.e., the attributes $C$) to control the image generation

The overview of Att-Adapter is provided in Fig.~\ref{fig:AttAdapter overview} (a). We use the pretrained image encoder, such as CLIP~\cite{radford2021learning} or a face recognition model~\cite{insightface,deng2018arcface,deng2019retinaface}. Inspired by \cite{sohn2015learning}, our CVAE module includes a variational encoder $q_\phi(Z|V,C)$, a conditional prior network $p_\psi(Z|C)$, and a decoder $p_\psi(V'|Z)$. We empirically remove the condition $C$ from the decoder, since the conditional prior $Z$ already contains the information of $C$. Complete derivations for the CVAE ELBO can be found in~\cite{sohn2015learning}. 

Let $V$ and $V'$ be input and output of the CVAE module, where $V$ can be obtained from the image encoding branch $f$. The output $V'$ is fed into the denoising UNet through the decoupled cross attention module, similar to the simple version of Att-Adapter. 

In accordance with the general CVAE setup, we assume that the prior and the posterior networks follow a Multivariate Gaussian which means both the prior and posterior networks estimate the Gaussian parameters $\mu$ and $\sigma$. As opposed to~\cite{zhao2017learning}, we assume an isotropic Gaussian only for the prior but not for the posterior as it is unusual in the image domain in practice, i.e., $q_\phi(Z|V,C) \sim \mathcal{N}(\mu, \Sigma)$, where $\Sigma=\text{diag}(\sigma_1^2,\sigma_2^2,...)$ and $p_\psi(Z|C) \sim \mathcal{N}(\mu, \sigma^2 \mathbf{I})$.

During the training, our variational encoder $q_\phi(Z|V,C)$ takes the multi-attributes condition with the image feature and approximate the gaussian parameters $\mu_q$ and $\sigma_q$ for the given condition. Our decoder $p_\psi(V'|Z)$ then takes the reparameterized sample and outputs $V'$. The conditional prior network $p_\psi(Z|C)$ is trained together and learns to produce conditional prior distributions $\mu_p$ and $\sigma_p$. The evidence lower bound  (ELBO) of the CVAE is defined as:
\begin{align} \label{eq:elbo}
    & \text{ELBO} \triangleq \nonumber \\
    & -\text{KL}\left( q_\phi(z|v,c) \, || \, p_\psi(z|c) \right) + \mathbb{E}_{q_\phi(z|v,c)} \left[ \log p_\psi(v'|z,c) \right] \nonumber \\
    & \leq \log \, p(V=v|C=c) .\nonumber\\
\end{align}
%where $\mathcal{L}_{\text{CVAE}}(v,c)=\text{-ELBO}$. 

%\vspace{0.1in}
\noindent \textbf{Training Objectives.}
%After we learn the conditional attributes via the CVAE branch, 
In conjunction with the CVAE objective, the standard denoising objective is applied, i.e., 
%derived by optimizing the variational bound of log likelihood  
$\mathcal{L}_{\text{denoising}}(x_t,t,y,c,x')=\mathbb{E} \left[ \| \epsilon - \epsilon_\theta(x_t,t,y,c,x') \|^2 \right].$ 
The final objective for optimizing our proposed Att-Adapter is defined as:
\begin{align}
    \mathcal{L}&(x_t,t,c,y,x') = \nonumber \\ &\mathcal{L}_{\text{CVAE}}(f(x'), c; \alpha) + \mathcal{L}_{\text{denoising}}(x_t,t,y,c,x'),
\label{eq:final objective}
\end{align}
where $\mathcal{L}_{\text{CVAE}}(f(x'),c; \alpha)$ is negative ELBO in Eq. \ref{eq:elbo}, where $\alpha$ is a coefficient for the KL term, and $f$ is a chain of functions to extract the image feature $v$ as shown in Fig.~\ref{fig:AttAdapter overview} (b). The full set of parameters to be updated is colored yellow in the figure while the green parts are frozen. We empirically use $\alpha=0.0001$.

Unlike the simple version of Att-Adapter that is trained only with the denoising objective, Att-Adapter with CVAE is designed to be robust against overfitting by minimizing $\mathcal{L}_{\text{CVAE}}$ together with the donoising loss. 

%represents the objective function of the domain-specific attributes $C$ as shown in Fig.~\ref{fig:graphical visualizations of ipadapter and attadapter}. On the other hand, $\mathcal{L}_{\text{denoising}}$ plays a role in fitting the domain-specific knowledge (from the edge $C \rightarrow X$) to the pretrained knowledge. Fig.~\ref{fig:AttAdapter overview} illustrates the full set of parameters to be updated during training. 

%\vspace{0.1in}
\paragraph{Reparameterization and Sampling.}
Following the CVAE literature~\cite{kingma2013auto}, we use the reparameterization trick for both training and sampling process. During training, reparameterization is applied for the posterior $q_\phi$ to make the sampling process differentiable.  During sampling, reparameterization is used to sample from the conditional prior distribution $p_\psi(Z|C)$ as shown in Fig.~\ref{fig:AttAdapter overview} (b).
%Let $g(\mu, \sigma, \epsilon)=\sigma\cdot \epsilon + \mu$ be a function for reparameterization. During training, reparameterization is applied for the posterior $q_\phi$ to make the sampling process differentiable, i.e., $g(\mu_q, \sigma_q, \epsilon)$, where $\mu_q$ and $\sigma_q$ are the approximated parameters from $q_\phi(Z|V,C)$. During sampling, reparameterization is used to sample from the conditional prior distribution $g(\mu_p, \sigma_p, \epsilon)$, where $\mu_p$ and $\sigma_p$ are from $p_\psi(Z|C)$.

% Following the CVAE literature~\cite{kingma2013auto}, we use the reparameterization trick for both training and sampling process. Let $g(\mu, \sigma, \epsilon)=\sigma\cdot \epsilon + \mu$ be a function for reparameterization. During training, reparameterization is applied for the posterior $q_\phi$ to make the sampling process differentiable, i.e., $g(\mu_q, \sigma_q, \epsilon)$, where $\mu_q$ and $\sigma_q$ are the approximated parameters from $q_\phi(Z|V,C)$. During sampling, reparameterization is used to sample from the conditional prior distribution $g(\mu_p, \sigma_p, \epsilon)$, where $\mu_p$ and $\sigma_p$ are from $p_\psi(Z|C)$.

%% file: sec/4_experiments.tex
\section{Experiments}
\label{sec:experiments}
More details on the data are in the appendix (Sec.~\ref{Appendix_sec:evaluation details}).
\subsection{Dataset and Preprocessing}
\label{subsec:data preprocessing}
\noindent\textbf{FFHQ}: FFHQ~\cite{karras2019style} is a high-quality public face dataset. To use the data for finetuning Text-to-image diffusion models, we get a caption per image by using ChatGPT~\cite{achiam2023gpt}. We preprocessed 20 facial attributes: 13 attributes for facial compositions, one attribute for age, and 6 attributes for race. Specifically, we first obtain head pose information (pitch, roll, and yaw)\footnote{https://github.com/DCGM/ffhq-features-dataset} and use them to get frontal view face images; We extract 16,336 images out of 70,000 images, which are used as our fine-tuning data. Next, we use 68 landmarks per image from dlib~\cite{king2009dlib} to obtain the information about facial composition. For example, `the gap between eyes' and  `width of mouth'. For age and race extraction, we use deepface~\cite{serengil2021lightface} with a detector RetinaFace~\cite{deng2019retinaface}.

\noindent\textbf{EVOX}: We used a commercial grade vehicle image dataset from EVOX\footnote{For more information, see \url{https://www.evoxstock.com/}.}. The dataset includes high-resolution images across multiple car models collected from 2018 to 2023, totally 2030 images.

%More details are described in the appendix.

% \footnote{The images used in this study are the property of EVOX Productions, LLC and are subject to copyright law. The appropriate licenses and permissions have been obtained to ensure the rightful use of these images in our study. For more information, see \url{https://www.evoxstock.com/}.} as the experiment data. The dataset includes high-resolution images across multiple car models collected from 2018 to 2023, totally 2030 images.

% \textbf{BIKED} Another dataset we use to apply our model is the BIKED Dataset \cite{biked}. This dataset is primarily collected for mechanical design applications, which includes both the parametric features and CAD files from 4500 bicycle designs. The parametric features for each full bicycle design is consisting of 222 features that include both discrete and continue values, such as $Saddle Height$, $Seat Tube Length$ and $Number of Bottles$.

\subsection{Evaluation Setup}
\label{subsec:evaluation setup}
This section defines our baselines and the metrics used for comparison. As shown in Table \ref{table: comparing the settings with baselines}, existing methods use two different settings for attribute control: \emph{absolute} and \emph{latent}. In the absolute setting, attribute values are explicitly provided using real-world data (e.g., absolute distances measured from images). In contrast, the latent setting assigns attribute values within a predefined concept space, such as those in GANs and CLIP, where values can only be manipulated within the given space.

\subsubsection{Baselines}

%\vspace{0.1in}
\hspace*{\parindent}\textbf{Baselines (Latent Control).}
Baseline methods for precise continuous attribute control require positive and negative paired data for training and often need a pretrained concept space ~\cite{li2024stylegan, gandikota2024concept,baumann2025attributecontrol} as discussed in Sec.~\ref{sec:related_works}. 

\noindent\textbf{S-I-GANs.} This baseline is implemented by combining StyleGAN 2~\cite{karras2019style} and InterFaceGAN~\cite{shen2020interfacegan}. We start from the 16,336 images from Sec.~\ref{subsec:data preprocessing} and their attribute labels. We first use the e4e~\cite{tov2021designing} encoder to project the real images onto the $W_+$ space. We then fit a SVM to get the semantic control direction per attribute by using the projected data and the corresponding labels. %We sample negative and positive pairs per attribute, where -7 and 7 are used for $\alpha$, which can be considered a strong modification according to InterFaceGAN. 

\noindent\textbf{$\mathbf{W_+}$ Adapter~\cite{li2024stylegan}.} The retrieved semantic directions from InterFaceGAN are used to get the negative and positive samples to evaluate.   % -4 and 4 are used for the scales which are exactly the default value by the original implementation.

\noindent\textbf{ConceptSlider~\cite{gandikota2024concept}.} The LoRA-based controllers from ConceptSlider require paired data for training. Since we do not usually have paired data for domain-specific attributes, e.g., face, we use the generated synthetic paired data from S-I-GANs to train ConceptSlider. We also train ConceptSlider (unpair) without paired data. Briefly, 1000 positive and another 1000 negative unpaired images are used. % -1.5 and 1.5 are used for the scales, following the original code repo. 

\noindent\textbf{AttrCtrl.~\cite{baumann2025attributecontrol}}
They do not use any image data to train their controller, so we make a text pair following the instructions. For example, the prompt target is set as `man' and the positive and the negative prompts are `man with big gap between eyes' and `man with small gap between eyes'.

%\vspace{0.1in}
\textbf{Baselines (Absolute Control).}
Next, we compare our performance with baseline methods that share the same data setting as ours; unpaired multi-attribute data.

\noindent\textbf{LoRA~\cite{hu2022lora}.}
Since ConceptSliders, which are based on LoRA, are not designed for training with unpaired data, we introduce another baseline that natively supports unpaired multi-attributes training. This approach first discretize continuous values and finetune using special tokens. For example, if two attributes have values of 0.2 and 0.4, the LoRA's condition is represented as `Z2 Y4'. Fig.~\ref{supfig:input settings of LoRA baseline} in the appendix provides a detailed example. To achieve optimal results, we apply LoRAs to both UNet and text encoder.  

\noindent\textbf{ITI-GEN~\cite{zhang2023iti}.}
ITI-GEN is not scalable for handling multiple conditions with numerous categories. Our goal is to model a conditional distribution that takes 20 attributes along with a text prompt, denoted as $p(x|y,c_1,c_2,...,c_{20})$. To achieve this within the ITI-GEN framework, we first generate inclusive prompts separately for every pair of attributes and model distributions such as $p(x|y)$, $p(x|c_1,c_2)$, ..., and $p(x|c_{19},c_{20})$. We then approximate the full conditional by merging the prompt information by averaging in the prompt embedding space (naive) or by leveraging Composable Diffusion Models (CDM)~\cite{liu2022compositional} in the score space.

% \vspace{0.1in}
\subsubsection{Measures} 

\hspace*{\parindent} \textbf{Baselines (Latent Control).} In this setting, attribute values come from the pretrained GANs' latent space rather than the actual facial attributes, making direct comparisons between input $c$ and predicted $\hat{c}$ (from the generated image) infeasible. Instead, we generate 200 paired images per attribute (e.g., closed vs. open eyes) and compare estimated attributes $\hat{c}_{neg}$ and $\hat{c}_{pos}$ using dlib~\cite{king2009dlib}. A higher score indicates a broader control range. We evaluate 13 individual attributes (2,600 images) and 10 multi-attribute combinations (2,000 images), applying a recommended or broader conditioning range than the original implementation (Sec.~\ref{Appendix_sec:evaluation details} in the appendix).

We measure the Control Range (CR) as the L1 distance between the target attributes of $\hat{c}_{neg}$ and $\hat{c}_{pos}$, where a higher value indicates better expressivity of the model with respect to the target attribute. Disentanglement Performance (DIS) is evaluated as the proportion of changes in the target attribute relative to total changes, with higher values indicating better disentanglement.

%\vspace{0.1in}
\textbf{Baselines (Absolute Control).} 
Baselines in this setting take a real unpaired condition $c$, allowing direct performance evaluation by comparing input $c$ with estimated $\hat{c}$ from the generated image using dlib~\cite{king2009dlib}. The lower error indicates better alignment between the generated images and the control attribute values.

We also measure the accuracy of race and age using deepface~\cite{serengil2021lightface} and Retinaface~\cite{deng2019retinaface}. The estimated age is compared with the input age via L1 distance, while race is evaluated using cross-entropy (lower is better).

Next, we measure the CLIP score~\cite{radford2021learning} to see if the generated images keep the pretrained knowledge. Lower scores suggest knowledge loss. %We compare the semantic similarity between the generated image and the text prompt. %If the pretrained knowledge is forgotten, the generated images might have worse (lower) similarity with the prompt embedding. 
We also use ChatGPT to verify alignment with the query ``Does the prompt `[PROMPT]' correctly describe the image? Shortly answer yes or no.''

For evaluation, we predefine 30 text prompts representing pretrained knowledge (which rarely/never exist in the finetuning data), e.g., `A marble sculpture of a man/woman'. For each prompt, we randomly sample 50 combinations of 20 attributes, totaling 1,500 test samples. Details can be found in Sec.~\ref{Appendix_sec:evaluation details} in the appendix.

For EVOX data, ChatGPT assesses whether the model correctly applies pose and body attributes to generated car. we use SegFormer~\cite{xie2021segformer} to measure the size accuracy. Briefly, we generate a pair of cars by giving the same random noise but different size conditions. Segmentation mask are then compared to quantify size differences.

% \vspace{0.1in}
% \subsubsection{Measures.}
% We evaluate Att-Adapter from two perspectives: (1) the semantic relevance with pretrained knowledge and (2) the effectiveness of implanting domain-specific knowledge for fine-grained attribute control.
% %We have got two aims in this paper. First, avoiding catastrophic forgetting and second, implanting the domain-specific knowledge.

% For FFHQ, we first measure if the generated images maintain the knowledge that is pretrained but not/rarely finetuned. To measure this, we first use CLIP~\cite{radford2021learning} score. We compare the semantic similarity between the generated image and the text prompt. If the pretrained knowledge is forgotten, the generated images might have worse (lower) similarity with the prompt embedding. We also use ChatGPT with a query of ``Does the prompt `[PROMPT]' correctly describe the image? Shortly answer yes or no.''

% We next measure if the facial attributes from the generated images are controlled well by the given conditions. We first use dlib~\cite{king2009dlib} to obtain the landmark. After obtaining, we can get the facial compositional attributes with which we can easily compute the L1 distance with input conditions. The lower is the better. We next use deepface~\cite{serengil2021lightface} and Retinaface~\cite{deng2019retinaface} to estimate the race and age of the generated images. The estimated age is compared with the input age in L1. The estimated race is compared with the input race conditions in cross entropy (lower is better).

\subsection{Evaluations}
%In this section, we thoroughly compare our performance with the baselines. 
\subsubsection{Baselines (Latent Control)}
\noindent We measure how well the target attribute is applied to the generated images in both single and multiple attributes. As shown in Table~\ref{table:baseline comparisons1, facial components difference}, our results outperform the baselines in both CR and DIS in the single and multiple attribute settings. The better disentanglment (DIS) scores of ours can be seen as a benefit of multi-attributes training. That is, all attributes are trained together learning the correlation between the attributes. This indicates that Att-Adapter can learn each attribute more precisely, which is also verified as the better CR score of ours. This is not surprising because the bigger the facial expressions are, the correlation between the attributes get bigger which is harder to learn without considering other attributes together.

The better CR score is also attributed to the fact that Att-Adapter can be trained with the real image and the real attributes as opposed to the baselines. Recall that the baselines require the pretrained GANs' space which essentially constraints their performance, i.e., both $W_+$Adapter and ConceptSliders are worse than S-I-GANs. This makes sense as the training dataset of $W_+$Adapter and ConceptSliders are the synthetic paired images generated by S-I-GANs.

Note that AttrCtrl performs not well in our evaluation protocols, as the method is not well-suited for visual attributes that cannot be easily described in text.

\begin{figure}[htp]
    \centering
    % \vspace{-0.1in}
    \includegraphics[width=1.0\linewidth]{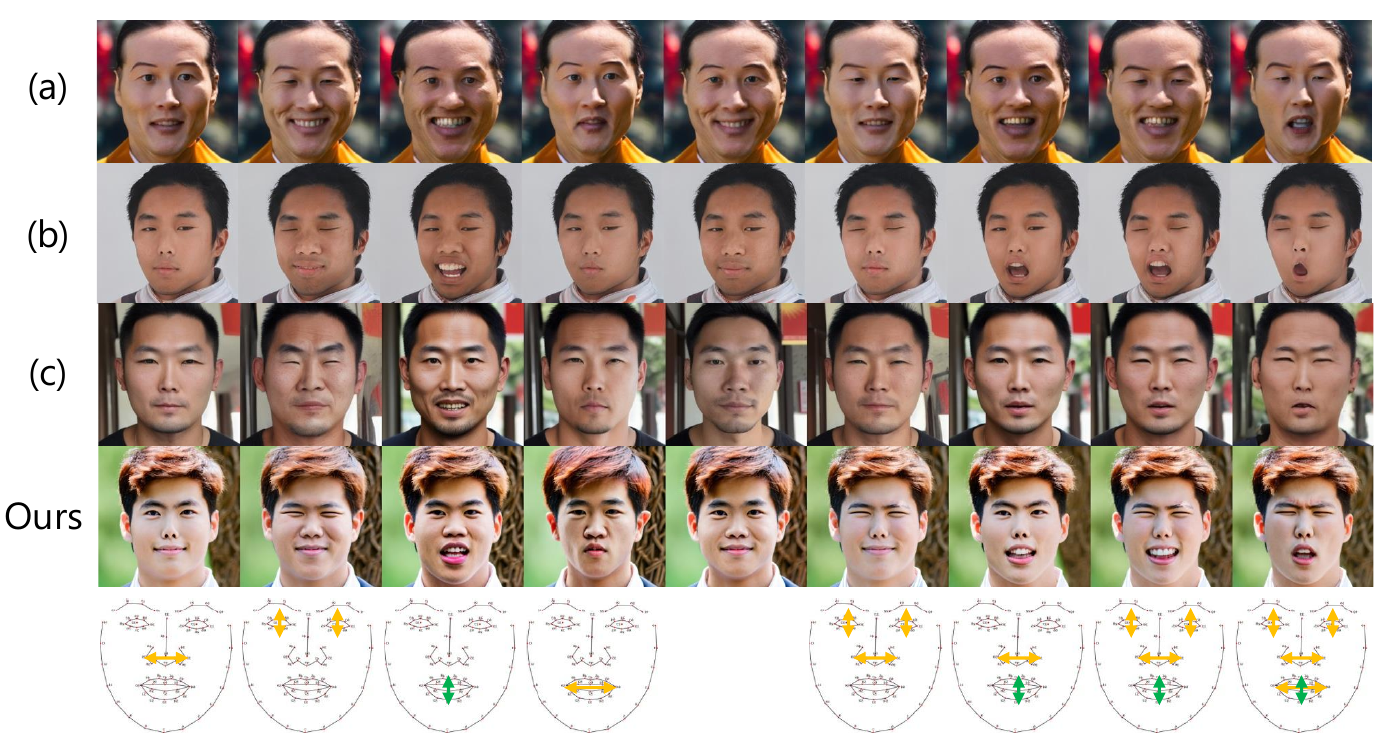}
    % \vspace{-0.1in}
    \caption{Best performing Latent Control Baselines; (a) ${W_+}$Adapter (b) S-I-GANs (c) ConceptSlider. Ours shows better performance in applying the target attribute with minimal change for the rest. The green and yellow arrows respectively indicate positive and negative direction controls.}
    \label{fig:baseline comparison}
\end{figure}

% inherent disentangling capability of VAE~(c.f., $\beta$-VAE). On top of the better DIS, 
% Furthermore, the joint conditioning modeling of Att-Adapter can benefit the disentangling performance because each attribute is learned while considering other attributes. This indicates that Att-Adapter can learn each attribute more precisely, which is also verified as the better CR score of ours. This is not surprising because the bigger the facial expressions are, the correlation between the attributes get bigger which is harder to learn without considering other attributes together.
Fig.~\ref{fig:baseline comparison} shows the qualitative results align with Table \ref{table:baseline comparisons1, facial components difference}. Both $W_+$Adapter and ConceptSliders show that the person identity across column is affected (low DIS) while the target attribute effects are weaker than ours (low CR). S-I-GANs shows relatively good attribute control performance as shown in the stronger target attribute effects than other baselines. For example, the rightmost three columns show bigger mouth openness than $W_+$Adapter and ConceptSliders. This also reveals the inherent limitation of the them; the pretrained GANs' performance cascades/upperbounds. Ours, on the other hand, shows the best performance in applying the target attribute better while having minimal change for the remaining attributes.

\begin{table}[htp]
\vspace{-0.1in}
\resizebox{\columnwidth}{!}{%

\begin{tabular}{cccccccc}
                                 &        & \multirow{2}{*}{Ours} & \multicolumn{2}{c}{ConceptSliders} & \multirow{2}{*}{AttrCtrl} & \multirow{2}{*}{$W_+$Adapter} & \multirow{2}{*}{S-I-GANs} \\ \cline{4-5}
                                 &        &                       & Pair         & Unpair       &                          &                          &                          \\ \hline
\multirow{2}{*}{CR ($\uparrow$)}   & Single & \textbf{67.87}                 & 26.73        & 8.64         & 9                        & 26.59                    & 49.45                    \\
                                 & Multi  & \textbf{90.39}                 & 31.35        & 10.5         & 20.78                    & 40.76                    & 68.4                     \\ \hline
\multirow{2}{*}{DIS ($\uparrow$)} & Single & \textbf{29.8\%}                & 17.5\%       & 10.4\%       & 7.6\%                    & 15.8\%                   & 21.8\%                   \\
                                 & Multi  & \textbf{32.3\%}                & 19.8\%       & 12.2\%       & 11.4\%                   & 18.5\%                   & 20.1\%                  
\end{tabular}
}
% \vspace{-0.1in}
\caption{Latent Control Baselines. CR indicates the control range of the attribute that each model is capable of, and DIS denotes disentangling performance. Our results outperform the baselines in both measures.}
\label{table:baseline comparisons1, facial components difference}
% \vspace{-0.15in}
\end{table}

\subsubsection{Baselines (Absolute Control)}

\begin{figure}[t]
  % \vspace{-0.2in}
  % \vspace{-0.1in}
  \centering
    \includegraphics[width=.5\textwidth]{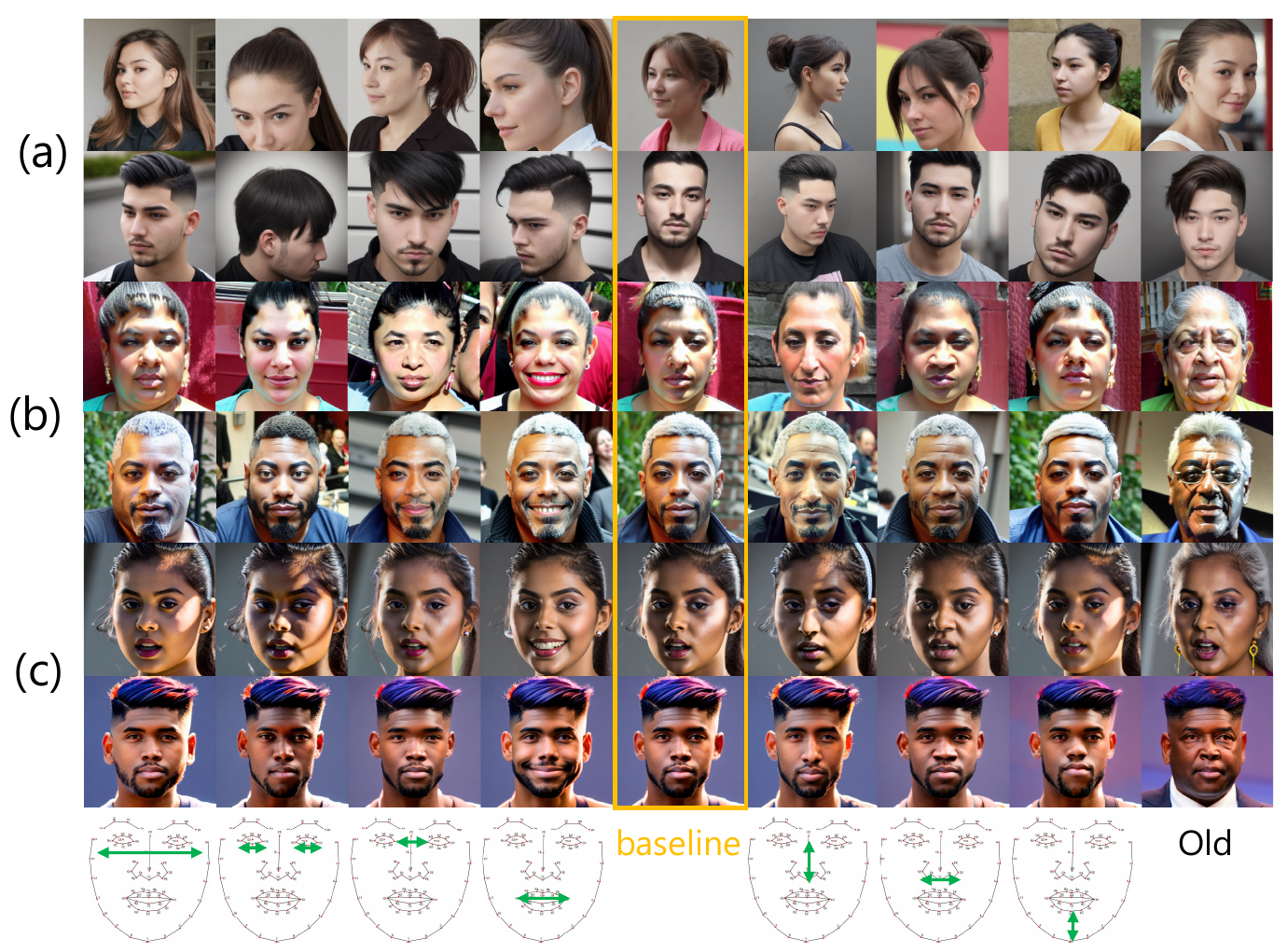}
  \caption{Qualitative results using Absolute Control baselines. From the top, (a) ITI-GEN, (b) LoRA, and (c) Att-Adapter.} %The prompts of ``A woman with ponytail'' and ``A man with shadow fade hair style'' are used for the woman images and the man images, respectively.} %`indian' and `black' are used for the woman and the man images, respectively.}
  \label{fig:baseline comparison on facial composition attributes}
  % \vspace{-0.15in}
\end{figure}

Fig.~\ref{fig:baseline comparison on facial composition attributes} shows the performance of ours and the baselines. Compared to the baseline column in the yellow box in the center, only one attribute is changed for each column. The changed target attribute is presented in the bottommost row. 

\begin{table}[]
% \vspace{-0.1in}
\begin{adjustbox}{width=\columnwidth,center}
\begin{tabular}{cccccc}

     & \multicolumn{3}{c}{Finetuned knowledge (L1$\downarrow$)}              & \multicolumn{2}{c}{Pretrained knowledge ($\uparrow$)} \\ \cline{2-6} 
     & Facial comp.   & Age          & Race           & CLIP                   & ChatGPT         \\ \hline
% (A)  & 69.43               &  8.8            &   1.3             &   \underline{0.298}                     &  \underline{97\%}               \\ \cline{1-6}
% (B)  & 18.89          & 10.0         & 0.64           & 0.271                  &  72\%               \\
(A)  & 17.48          & 10.0         & 0.48           & 0.270                  & 72\%            \\
% (D)  & 16.11          & 10.4         & 1.74           & 0.271                  &      73\%           \\
(B)  & 52.16          & 8.8          & 1.77           & \textbf{0.288}         &       79\%          \\
(C)  & 70.09          & 9.3          & 2.06           & 0.286                  & \textbf{84\%}            \\
Ours & \textbf{15.57} & \textbf{8.0} & \textbf{0.36} & 0.283                  & 78\%           
\end{tabular}
\end{adjustbox}
\caption{Quantitative results using Absolute Control baselines. (A) LoRA, (B) ITI-GEN~\cite{zhang2023iti} (naive), (C) ITI-GEN (CDM~\cite{liu2022compositional}).}
\label{table: baseline comparisons ffhq}
\vspace{-0.1in}
\end{table}
% \begin{table}[]
% \begin{adjustbox}{width=\columnwidth,center}
% \begin{tabular}{cccccc}

%      & \multicolumn{3}{c}{Finetuned knowledge ($\downarrow$)}              & \multicolumn{2}{c}{Pretrained knowledge ($\uparrow$)} \\ \cline{2-6} 
%      & Facial comp.   & Age          & Race           & CLIP                   & ChatGPT         \\ \hline
% (A)  & 69.43               &  8.8            &   1.3             &   \underline{0.298}                     &  \underline{97\%}               \\ \cline{1-6}
% (B)  & 18.89          & 10.0         & 0.64           & 0.271                  &  72\%               \\
% (C)  & 17.48          & 10.0         & 0.48           & 0.270                  & 72\%            \\
% (D)  & 16.11          & 10.4         & 1.74           & 0.271                  &      73\%           \\
% (E)  & 52.16          & 8.8          & 1.77           & \textbf{0.288}         &       79\%          \\
% (F)  & 70.09          & 9.3          & 2.06           & 0.286                  & \textbf{84\%}            \\
% Ours & \textbf{15.57} & \textbf{8.0} & \textbf{0.36} & 0.283                  & 78\%           
% \end{tabular}
% \end{adjustbox}
% \caption{Quantitative baseline comparisons. (A) without finetuning, (B) LoRA~\cite{hu2022lora} with 128 dimension, (C) LoRA d512, (D) LoRA d2048, (E) ITI-GEN~\cite{zhang2023iti} (naive), (F) ITI-GEN (CDM~\cite{liu2022compositional}).}
% \label{table: baseline comparisons ffhq}
% \vspace{-0.1in}
% \end{table}

\vspace{0.1in}
Fig.~\ref{fig:baseline comparison on facial composition attributes} (a) ITI-GEN does not show meaningful semantic changes for both facial compositing attributes and age. This is because ITI-GEN is limited in combining multiple conditions post hoc, as described in details in sec.~\ref{subsec:evaluation setup}. It is also shown in Table~\ref{table: baseline comparisons ffhq} (B) and (C). Both naive and cdm methods are not correctly applying the finetuned knowledge to the generation process, even though it maintains the pretrained knowledge well. This is because ITI-GEN is not scalable to many attributes by design (e.g., 20 in our experiments)

%essentially has too many separately trained attributes embeddings in our experiment setting, e.g., $p(x|c_1,c_2)$, $p(x|c_3,c_4)$, ..., $p(x|c_{19},c_{20})$ to approximate the whole conditional $p(x|y,c_1,c_2,...,c_{20})$. We believe the attribute embeddings obtained under ITI-GEN framework are getting diluted by mingling with other multiple (more than ten) attributes. Note that the scalability is one of the fundamental limitation of ITI-GEN. 

\vspace{0.1in}
\noindent\textbf{LoRA v.s. Ours in FFHQ.}
Fig.~\ref{fig:baseline comparison on facial composition attributes} (b) and (c) show the results of LoRA and ours. The results from both methods show decent performance in applying the target attribute while maintaining the original knowledge. However, Att-Adapter outperforms LoRA in many aspects. 

\begin{figure}[h]
  \vspace{-0.1in}
  \centering
    \includegraphics[width=.5\textwidth]{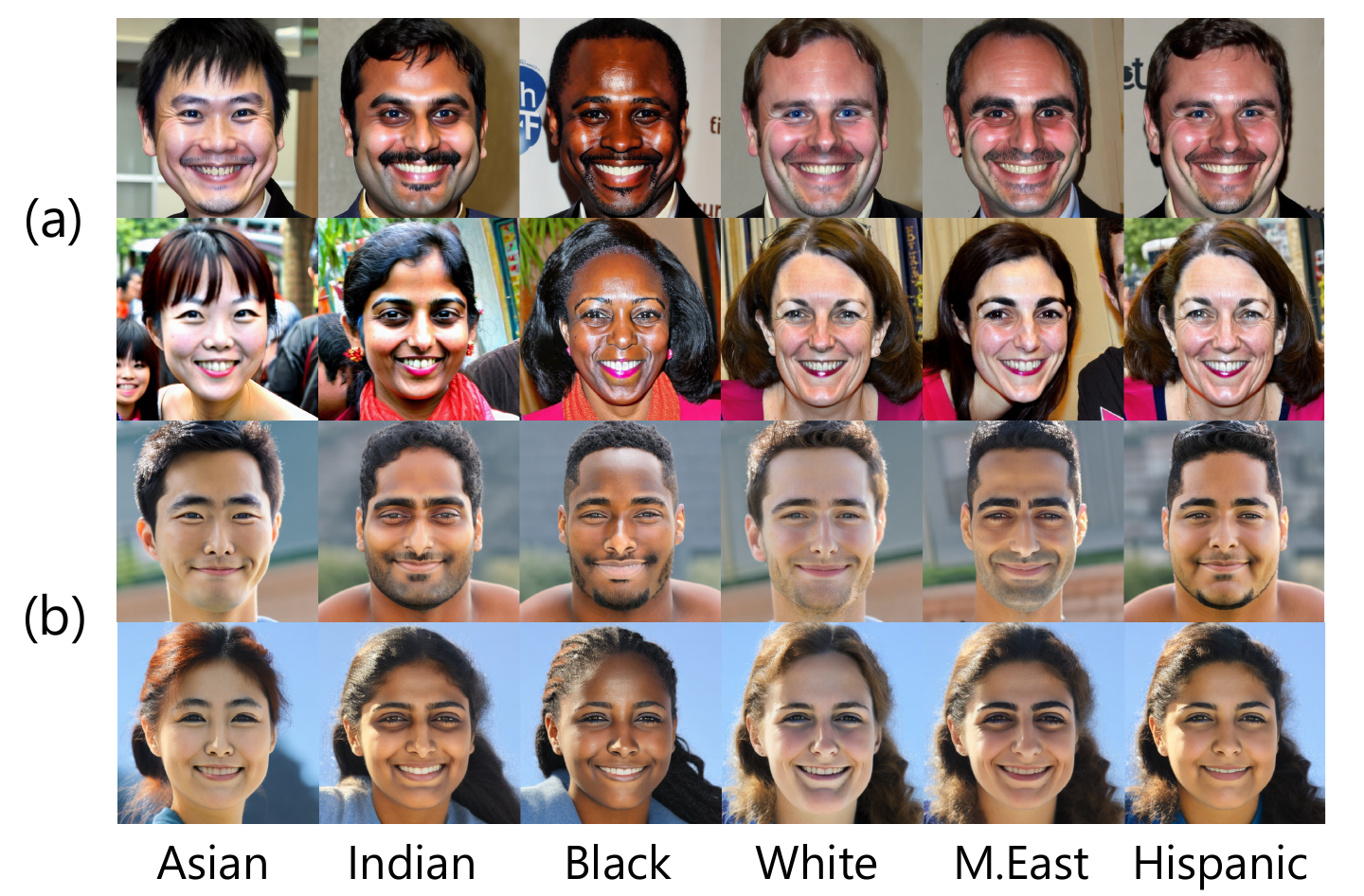}
  \caption{Qualitative Baseline comparisons on race. From the top, (a) LoRA, and (b) Att-Adapter. The prompts of ``A photo of a woman smiling'' and ``A photo of a man with shadow fade hair style'' are used for the woman images and the man images.}
  \label{supfig:baseline comparison on race}
  \vspace{-0.1in}  
\end{figure}

First of all, their visual fidelity is not as good as ours in (c). This is because Att-Adapter can naturally combine two different conditions (i.e., text and the attributes) by leveraging the decoupling cross-attention layer and the strong embedding power of the pretrained image encoder. Secondly, the performance for applying the facial attributes is worse than ours. For example, in the row (b) in the sixth column from the left, the height of nose changes a lot of other visual features together such as face length and mouth shape, while ours does less. Similarly, in Table~\ref{table: baseline comparisons ffhq}, ours shows better performance in correctly applying the given facial component attributes. Third, our Att-Adapter shows better performance in correctly applying the given race attribute as shown in Table~\ref{table: baseline comparisons ffhq} by the lower cross entropy. The better performance of Att-Adapter over the baseline in the race attribute can be found in Fig.~\ref{supfig:baseline comparison on race}. 
This can be observed by comparing the fourth and sixth columns of (a), which shows that LoRA is confused with the generation of white and Hispanic.

We believe this is caused by the fundamental limitation of discretization. It shrinks the amount of the original information which can cause distortions of the correlation between the facial features. 
%Even though it would not reverse the correlational information, the distortion can cause the model to mistake the attribute and to map it to the wrong cluster of the distorted visual features during the training. Since the facial attributes are very subtle domain, the small changes can yield completely different results, not in the ideal way.
Furthermore, discretizing the value is inherently limited in extrapolating the attributes. Please see Fig.~\ref{supfig:extrapolating example} in the appendix, which shows the advantage of our method over the LoRA baseline. 

\begin{figure}[t]
  % \vspace{-0.1in}
  \centering
    \includegraphics[width=.5\textwidth]{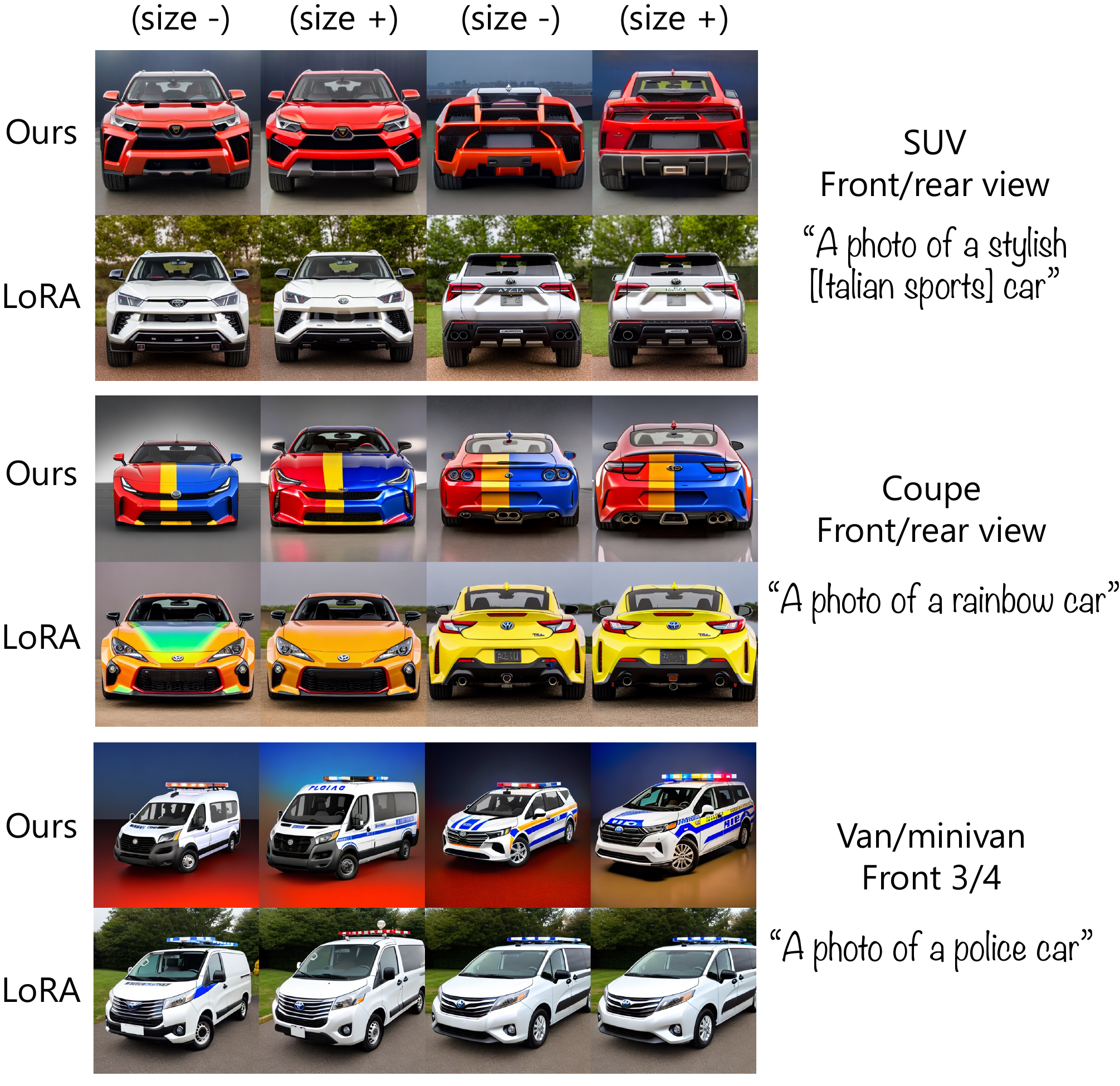}

  \caption{Absolute control baseline comparisons in EVOX dataset. [sports brand] refers to a well-known luxury sports car manufacturer.}
  \label{fig:baseline comparison in evox}
  \vspace{-0.1in}
\end{figure}

%Note that even each component $p(x|c_1,c_2)$ has 100 embeddings (e.g., assuming $c_1$ and $c_2$ are `width of nose' and `height of nose',  $c_1=\{0,1,...,9\}$ and $c_2=\{0,1,...,9\}$ 
\noindent\textbf{LoRA v.s. Ours in EVOX.}
To demonstrate the generalizability of Att-Adapter beyond human face dataset, we conducted an experiment on EVOX dataset. Fig.~\ref{fig:baseline comparison in evox}, shows the results. First, both Att-Adapter and LoRA generate well the discrete attributes such as pose, e.g., front view, and body of the car, e.g., SUV. For example, on the first macro row, both models consistently generate SUVs. However, we can see that ours show better performance in controlling the continuous attribute like size of the car, compared to LoRA. Moreover, we can see that Att-Adapter shows better performance in maintaining the pretrained knowledge. For example, given the prompt of ``A photo of a rainbow car'', the results from ours show stronger prompt effects.

These observations can be verified by Table~\ref{table: baseline comparison evox results} quantitatively. While the performance of LoRA and Att-Adapter is comparable in Pose and Body, Att-Adapter outperforms the baseline in continuous attributes and maintaining the original knowledge, which is consistent with FFHQ results.

% \begin{figure}[t]
%   \vspace{-0.2in}
%   \centering
%     \includegraphics[width=.5\textwidth]{figures/main_comparison_race.pdf}
%   \caption{Qualitative Baseline comparisons on race. From the top, (a) LoRA, and (b) Att-Adapter. The prompts of ``A photo of a woman with smiling'' and ``A photo of a man with shadow fade hair style'' are used for the woman images and the man images.}
%   \label{fig:baseline comparison on race}
%   \vspace{-0.1in}
% \end{figure}

% \begin{figure}[t]
%   % \vspace{-0.2in}
%   \centering
%     \includegraphics[width=.5\textwidth]{figures/extrapolation.pdf}
%   \caption{Extrapolation comparisons with LoRA showing the strength of Att-Adapter. A prompt of ``A stylish man smiling'' is used with `Asian' condition.}
%   \label{fig:extrapolating example}
%   \vspace{-0.1in}
% \end{figure}

\begin{table}[h]
% \vspace{-0.1in}
\begin{adjustbox}{width=.75\columnwidth,center}
\begin{tabular}{ccccc}
            & Pose ($\uparrow$) & Body ($\uparrow$) & Size ($\uparrow$) & CLIP ($\uparrow$) \\ \hline
% w.o. tuning & 97\% & 68\% & 52\% & 0.22 \\
LoRA        & \textbf{99\%} & 75\% & 67\% & 0.22 \\
ours        & 98\% & \textbf{76\%} & \textbf{95\%} & \textbf{0.24}
\end{tabular}
\end{adjustbox}
\caption{Absolute control baseline comparisons in EVOX dataset.}
\label{table: baseline comparison evox results}
\vspace{-0.15in}
\end{table}

\subsection{Analysis and Application}
%In this section, analysis on Att-Adapter is provided.

\begin{figure}
    \centering
    % \vspace{-0.1in}
    \includegraphics[width=0.8\linewidth]{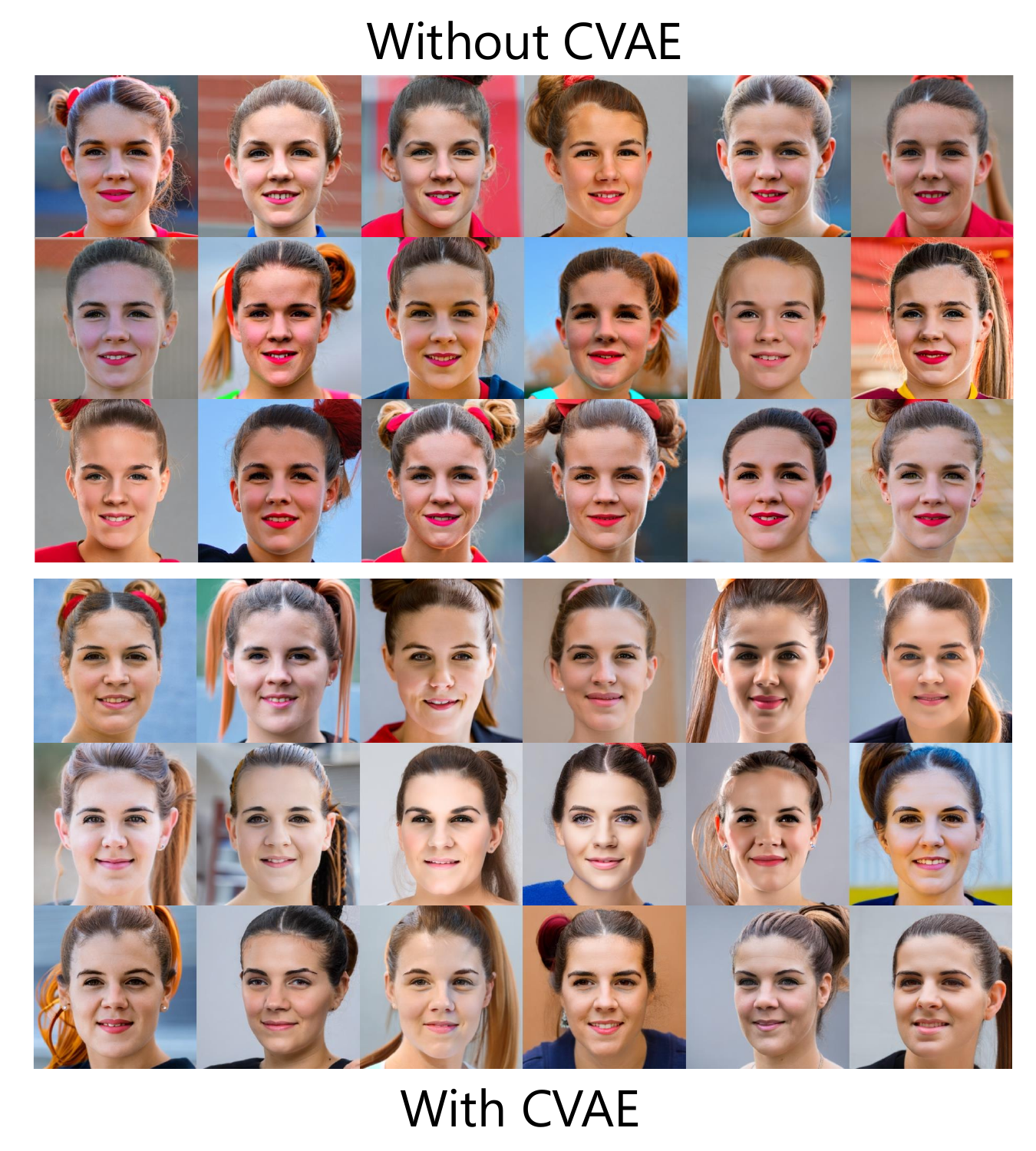}
    \vspace{-0.1in}
    \caption{Overfitting phenomenon. All of the conditions such as the attributes and the text prompt are fixed while the random sample such as $x_T$ and $\epsilon \sim N(0,I)$ for reparameterization are changed. It is not surprising that the generated images look similar as the given $c$ is fixed. However, the fixed $c$ does not necessarily mean the fixed person ID. The output person's identity is saturated without CVAE setting, which is not ideal as shown in Table~\ref{table:ablation study without cvae}.}
    % \vspace{-0.1in}
    \label{fig:random sampling showing overfitting}
\end{figure}

\vspace{0.1in}
\noindent\textbf{Att-Adapter with and without CVAE.} 
%Our motivation is to leverage CVAE's ability to learn a latent space that can accommodate both categorical and continuous variables over multiple attributes. 
We conducted additional ablation studies that compare the CVAE with an alternative MLP. Fig.~\ref{fig:random sampling showing overfitting} shows that using CVAE naturally regularizes the Adapter from overfitting. The results without CVAE shows low image diversity and convergence to similar facial appearances. Consistent patterns are observed in Table \ref{table:ablation study without cvae}. Although more training iterations improve attribute control, the person ID diversity significantly decreases. 

\begin{table}[]
\vspace{-0.1in}
\centering
\resizebox{.85\columnwidth}{!}{%
\begin{tabular}{ccccc}
             & \multicolumn{3}{c}{w.o. CVAE} &     \\ \cline{2-4}
(Iter)    & (20k)    & (50k)   &  (200k)  & with CVAE (200k)   \\ \hline
CR ($\uparrow$) & 47.04   & 62.06  & 78.27  & 78.59 \\
ID Sim ($\downarrow$)      & 6.0\%    & 27.1\%  & 28.5\%  & 6.2\% 
\end{tabular}
}
\caption{ID Sim stands for the average face verification scores by DeepFace (Facenet512) between all the pairs of 100 randomly generated images while feeding a fixed attribute conditions $c$. The results show that CVAE regularizes Att-Adapter from overfitting.}
\label{table:ablation study without cvae}
% \vspace{-0.1in}
\end{table}

\begin{table}[]
% \vspace{-0.1in}
\begin{adjustbox}{width=\columnwidth,center}
\begin{tabular}{cccccc}
     & \multicolumn{3}{c}{Finetuned knowledge ($\downarrow$)}              & \multicolumn{2}{c}{Pretrained knowledge ($\uparrow$)} \\ \cline{2-6} 
($\lambda$.)   & Facial comp.   & Age          & Race           & CLIP                   & ChatGPT         \\ \hline

0.0  &     69.43           &      8.8        &       1.3         &  \textbf{0.298}                      &     \textbf{97\%}            \\

0.5  &     28.96           &      8.5        &       0.50         &  0.293                      &     94\%            \\
% 150k  & 17.53          & 7.7         & 0.98           & 0.282                  &                 \\
0.7  & 17.96          & \textbf{8.0}         & 0.38           &  0.288                 &     89\%        \\
% 250k  & 16.29          & 7.4         & 0.46           & 0.283                  &                 \\
0.9  & \textbf{15.57}          & \textbf{8.0}         & \textbf{0.36}           & 0.283                  & 78\%

\end{tabular}
\end{adjustbox}
\caption{Performance reports with different $\lambda$s. The lower the $\lambda$ is, the stronger the pretrained knowledge affects the generation process while relatively weakening the influence of Att-Adapter.}
\label{table: sampling performance per lambda}
\vspace{-0.1in}
\end{table}

\begin{figure}[t]
  
  \centering
    \includegraphics[width=.5\textwidth]{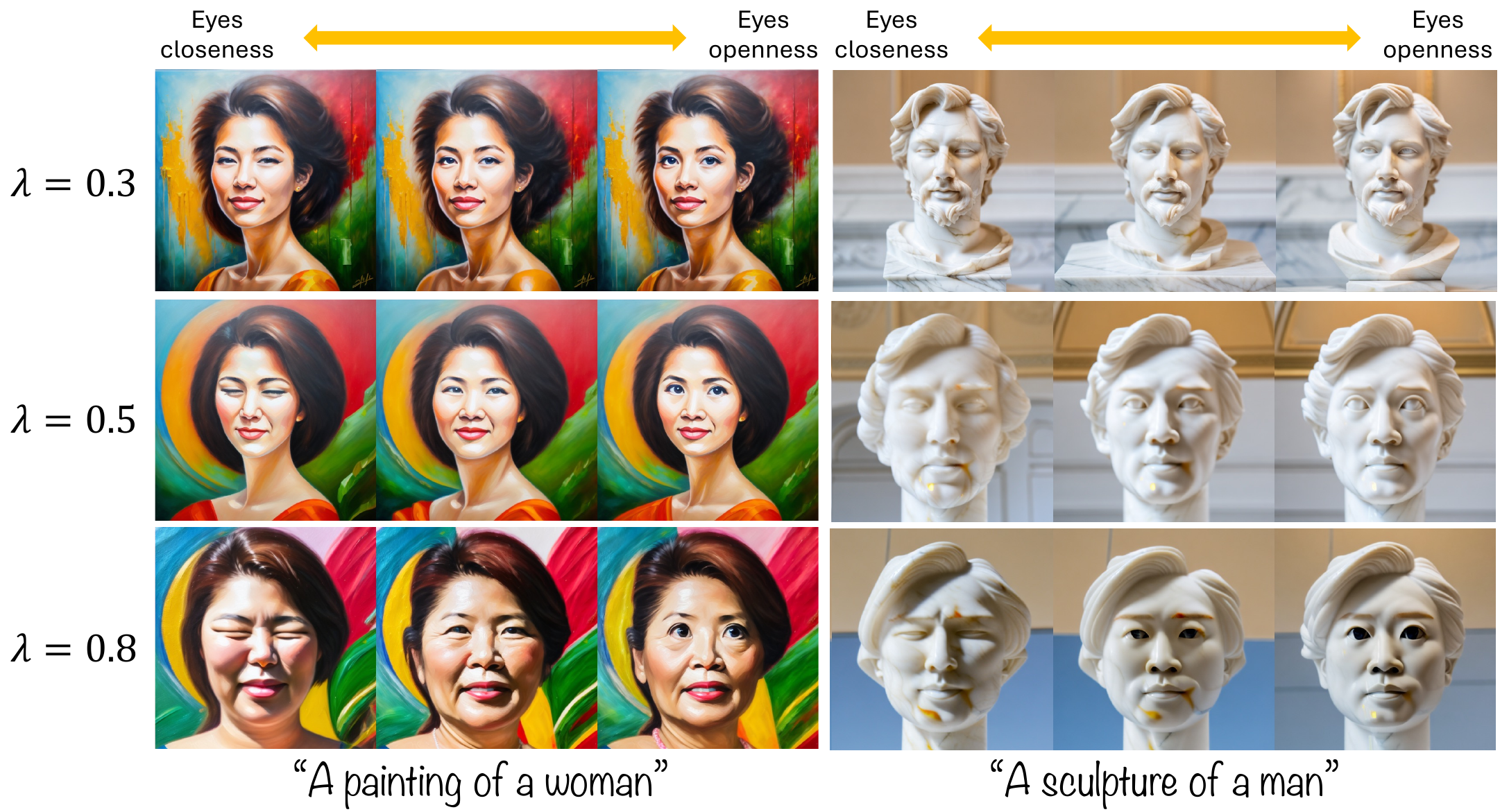}
  
  \caption{Qualitative explorations on the effects of $\lambda$.}
  \label{fig:lambda comparisons}
  \vspace{-0.1in}
    
\end{figure}

\begin{figure*}[t]
  % \vspace{-0.1in}
  \centering
    \includegraphics[width=.95\textwidth]{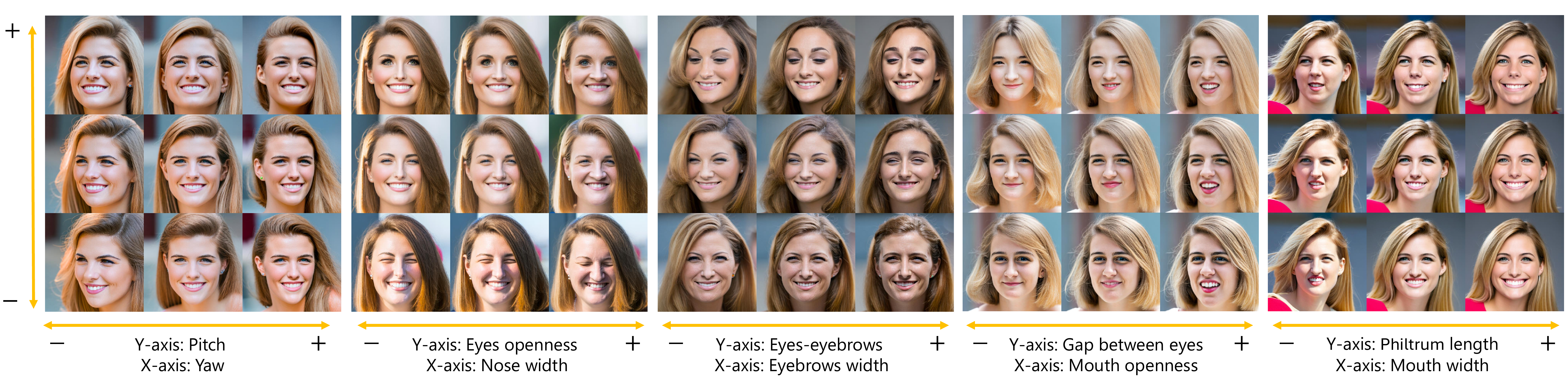}
    \vspace{-0.1in}
  \caption{Additional multi-attributes control examples. A prompt of ``A woman smiling'' is used to generate. Only a single Att-Adapter is used for all of the controlling examples.}
  \label{fig:multi-attributes control}
% \vspace{-0.15in}
\end{figure*}

\begin{figure}[t]
  % \vspace{-0.15in}
  % \vspace{-0.1in}
  \centering
    \includegraphics[width=.5\textwidth]{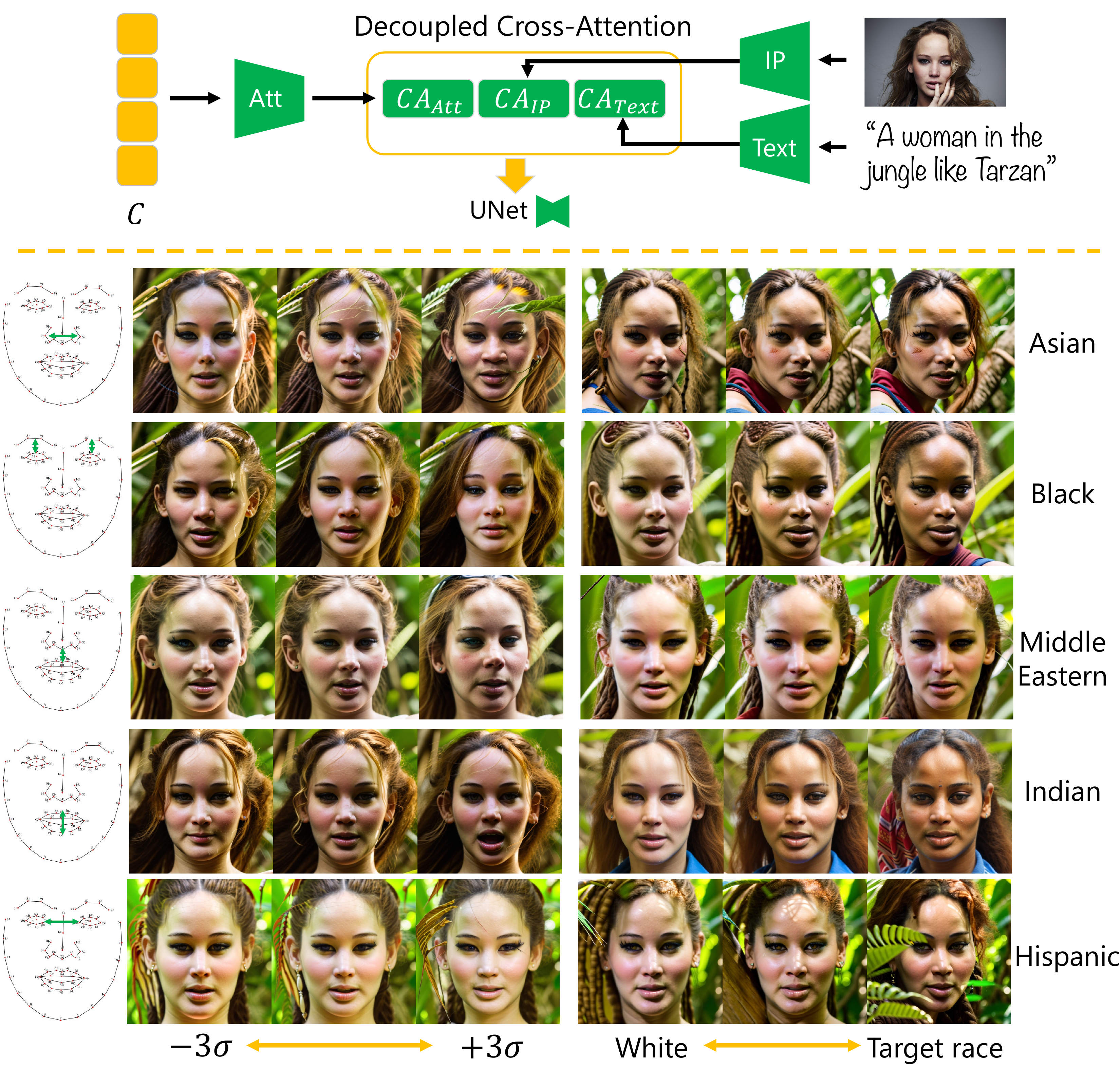}
  \caption{Examples on adapting both IP- and Att-Adapter to pretrained diffusion models to take an image as input. $\sigma$ is precomputed standard deviation per attribute from the finetuning data.}
  \label{fig:ipadapter_example}
  % \vspace{-0.2in}
\end{figure}

\vspace{0.1in}
\noindent\textbf{$\lambda$ during the sampling.} 
%As shown in Eq.~\ref{eq:decoupled cross attention}, $\lambda$ controls the strength of Att-Adapter when merging the information with text conditions. We show that Att-Adapter can be harmonized well with the pretrained diffusion models given varied $\lambda$s. (Fig.~\ref{fig:lambda comparisons} and Table~\ref{table: sampling performance per lambda} in the appendix.)
As shown in Eq.~\ref{eq:decoupled cross attention} in the main paper, $\lambda$ controls the strength of Att-Adapter when merging the information from different input conditions. Intuitively, by reducing $\lambda$, we can expect that Att-Adapter will affect the generation process less. This can be verified by looking at Fig.~\ref{fig:lambda comparisons}. When $\lambda=0.3$, we can see that the effect of text prompt is strong while the attribute effect is weak. For example, the sculptures does not look like the given attribute, `Asian'. When $\lambda=0.8$, the effect of Att-Adapter (where the finetuning knowledge passes through) gets stronger as we can see that the results in the third row have more human-like texture and face-cropped view. The quantitative results can be found in Table.~\ref{table: sampling performance per lambda}. We can first see that pretrained knowledge, measured by CLIP and ChatGPT is maintained better with lower $\lambda$. The higher the $\lambda$ is, on the other hand, we can expect the stronger effects of the fine-tuned knowledge, in exchange for losing the pretrained knowledge.

\vspace{0.1in}
\noindent\textbf{Exploring training settings.}
We explore several different settings of our method. The details are in Sec.~\ref{Appendix_sec:exploring training settings} in the appendix. Some of the takeaways from the experiments are as follows. 1) Model performance is converged around 200k iterations with batch size of 4. 2) Embedding dimension d (i.e., $Z\in\mathbb{R}^{\text{d}}$) needs to be bigger than 128 such as 512 or 2048. 3) The performance difference between isotropic and non-isotropic gaussian prior is marginal.

% \begin{table}[]
% \label{table: performance per iteraton}
% \begin{adjustbox}{width=\columnwidth,center}
% \begin{tabular}{cccccc}
%      & \multicolumn{3}{c}{Finetuned knowledge ($\downarrow$)}              & \multicolumn{2}{c}{Pretrained knowledge ($\uparrow$)} \\ \cline{2-6} 
% (iter)   & Facial comp.   & Age          & Race           & CLIP                   & ChatGPT         \\ \hline

% 100k  &     21.1           &      9.8        &       2.48         & \textbf{0.285}                       &    \textbf{82\%}             \\
% % 150k  & 17.53          & 7.7         & 0.98           & 0.282                  &                 \\
% 200k  & \textbf{15.57}          & 8.0         & \textbf{0.36}           & 0.283                  & 78\%            \\
% % 250k  & 16.29          & 7.4         & 0.46           & 0.283                  &                 \\
% 300k  & 15.92          & \textbf{7.7}          & \textbf{0.36}           & 0.282         &    77\%   
% \end{tabular}
% \end{adjustbox}
% \caption{Performance per iterations. We observe that the performance is empirically converged around 200k.}
% \end{table}

% \usepackage{multirow}

\vspace{0.1in}
\noindent\textbf{Controlling multiple attributes.}
Fig.~\ref{fig:baseline comparison} and Fig.~\ref{fig:multi-attributes control} shows the superior performance of Att-Adapter in controlling multi-attributes. More examples and a detailed explanation about the additional attributes yaw and pitch are provided in the appendix (Sec.~\ref{Appendix_subsec:challenging attributes}).

% \begin{figure}[h]
%   \vspace{-0.1in}
%   \centering
%     \includegraphics[width=.5\textwidth]{figures/yaw_pitch.pdf}
%   \caption{Qualitative results on additional attributes control. The results are generated by taking a prompt of ``A smiling woman''.}
%   \label{fig:yaw pitch results}
%     \vspace{-0.1in}
% \end{figure}

% \begin{figure}[h]
%   \vspace{-0.1in}
%   \centering
%     \includegraphics[width=.35\textwidth]{figures/yaw_eye.pdf}
%   \caption{Qualitative results showing that two attributes can be controlled simultaneously. The results are generated by taking a prompt of ``A smiling woman''.}
%   \label{fig:yaw eye at the same time}
%     \vspace{-0.1in}
% \end{figure}

\vspace{0.1in}
\noindent\textbf{Application: Augmenting Other Adapters.} To show more potential applications of Att-Adapter, we connect IP-Adapter\cite{ye2023ip-adapter} with Att-Adapter to take an image input while controlling the attributes of the input image by Att-Adapter. Details are shown in the top of Fig.~\ref{fig:ipadapter_example}. During the inference, we attach IP-Adapter and Att-Adapter together to pretrained Diffusion Models. The Decoupled Cross-Attention layer is extended to $\text{CA}(Q,K_{\text{text}},V_{\text{text}}) + \lambda_1 \text{CA}(Q,K_{\text{attr}},V_{\text{attr}}) + \lambda_2 \text{CA}(Q,K_{\text{ip}},V_{\text{ip}})$, where $\lambda_1$ and $\lambda_2$ determine the strength of Att-Adapter and IP-Adpater, respectively. We use 0.35 and 1.0 in practice. Interestingly, as shown in Fig.~\ref{fig:ipadapter_example}, the attribute information from Att-Adapter can be harmonized well with IP-Adapter, even though not trained together.  

%\vspace{0.1in}
%\noindent\textbf{Application 2: LoRA.} 
We also show the results of combining LoRA finetuned for a person ID with Att-Adapter in the appendix (Sec.~\ref{Appendix_sec:lora application}).

%% file: sec/5_conclusion.tex
\section{Conclusion and Discussion}
\label{sec:conclusion}
We present Att-Adapter for T2I diffusion models to enable fine-grained attribute control by adjusting continuous numeric values. Att-Adapter leverages decoupled cross-attention module to merge the conditions under different modalities. CVAE is further introduced to mitigate the overfitting issue. Different from the baselines, Att-Adapter can be trained without any paired data while can learn multi-attributes at once. The learned adapter allows meaningful and precise manipulation of attributes in the targeted domain. The experiment results demonstrate that Att-Adapter can more accurately control continuous attribute variables across multiple datasets compared to the baselines.
%\\\textbf{Limitations.} (1) Effectiveness across attributes: Many attributes are continuous variables instead of discrete categories. Att-Adapter provides more improvements for the attributes that require fine-grained adjustment along a continuum, but less improvements for the discrete and semantically describable attributes. (2) Additional labels: Att-Adapter needs annotations of the fine-grained attributes for training. While many attributes can take advantage of well established predictors for automatic annotation such as those demonstrated in our paper, some attributes may need extra manual efforts to obtain the annotations.
%(3) Generalizability: Att aims to be an adapter for generic T2I generative models. Though we choose a particular backbone in our current implementation, we will investigate the use with more T2I backbones in the future. 
\\\textbf{Potential Societal Impacts.} This study uses face dataset in the experiments for fair comparisons with the prior works. However, the approach is generic for other objects such as cars as demonstrated in the paper. We encourage a shift towards use cases in object manipulation and product design to reduce ethical concerns about privacy and discrimination.

% \vspace{0.1in}
% \textbf{Limitations.}

% \vspace{0.1in}
% \textbf{Societal Impacts.}

% \vspace{0.1in}
% \textbf{Conclusion.}

%% file: rebuttal.tex
% \documentclass[10pt,twocolumn,letterpaper]{article}
% \usepackage[rebuttal]{cvpr}

% % Include other packages here, before hyperref.
% \usepackage{graphicx}
% \usepackage{amsmath}
% \usepackage{amssymb}
% \usepackage{booktabs}

% % table
% \usepackage{adjustbox}
% \usepackage{multirow}

% % Import additional packages in the preamble file, before hyperref
% \input{preamble}

% % If you comment hyperref and then uncomment it, you should delete
% % egpaper.aux before re-running latex.  (Or just hit 'q' on the first latex
% % run, let it finish, and you should be clear).
% \definecolor{cvprblue}{rgb}{0.21,0.49,0.74}
% \usepackage[pagebackref,breaklinks,colorlinks,allcolors=cvprblue]{hyperref}

% % If you wish to avoid re-using figure, table, and equation numbers from
% % the main paper, please uncomment the following and change the numbers
% % appropriately.
% %\setcounter{figure}{2}
% %\setcounter{table}{1}
% %\setcounter{equation}{2}

% % If you wish to avoid re-using reference numbers from the main paper,
% % please uncomment the following and change the counter value to the
% % number of references you have in the main paper (here, 100).
% %\makeatletter
% %\apptocmd{\thebibliography}{\global\c@NAT@ctr 100\relax}{}{}
% %\makeatother

% %%%%%%%%% PAPER ID  - PLEASE UPDATE
% \def\paperID{16524} % *** Enter the Paper ID here
% \def\confName{CVPR}
% \def\confYear{2025}

% \begin{document}

% %%%%%%%%% TITLE - PLEASE UPDATE
% \title{Supplementary Material \\ \normalsize{Domain-Specific Att-Adapter Augmenting Pretrained T2I Diffusion Models}}  % **** Enter the paper title here

% \maketitle
\thispagestyle{empty}
\appendix

%%%%%%%%% BODY TEXT - ENTER YOUR RESPONSE BELOW
\section{Evaluation Details}
\label{Appendix_sec:evaluation details}

\paragraph{Dataset and Preprocessing}
\noindent\textbf{FFHQ}: FFHQ~\cite{karras2019style} is a high-quality public face dataset. To use the data for finetuning Text-to-image diffusion models, we get a caption per image by using ChatGPT~\cite{achiam2023gpt}. We preprocessed 20 facial attributes: 13 attributes for facial compositions, one attribute for age, and 6 attributes for race. 

Specifically, we first obtain head pose information (pitch, roll, and yaw)\footnote{https://github.com/DCGM/ffhq-features-dataset} and use them to get frontal view face images; We extract 16,336 images out of 70,000 images, which are used as our fine-tuning data. Next, we use 68 landmarks per image from dlib~\cite{king2009dlib} to obtain the information about facial composition. For example, to obtain `the gap between eyes', we use the 39th and 42nd landmarks coordinates and compute the L1 norm. Min-max normalization is applied to make the range of the value within 0 to 1, which means that 1 and 0 respectively indicate the max and min values of a certain attribute from the fine-tuning dataset. The 13 attributes related to facial compositions are: `the gap between eyes', `width of eyes', `height of eyes', `width of nose', `height of nose', `width of mouth', `height of mouth', `width of face', `height of face' (from glabella to chin), `height between eyebrow and eye', `height between nose and mouth', `height between mouth and chin', and `width of eyebrow'.

For age and race extraction, we use deepface~\cite{serengil2021lightface} with a detector Retinaface~\cite{deng2019retinaface}. The age is divided by 100 for normalization purposes. The six extracted features about the race are `p(Asian)', `p(Black)', `p(Hispanic Latino)', `p(Indian)', `p(Middle Eastern)', `p(White)'. 

\noindent\textbf{EVOX}: We used a commercial grade vehicle image dataset from EVOX \footnote{The images used in this study are the property of EVOX Productions, LLC and are subject to copyright law. The appropriate licenses and permissions have been obtained to ensure the rightful use of these images in our study. For more information, see \url{https://www.evoxstock.com/}.} as the experiment data. The dataset includes high-resolution images across multiple car models collected from 2018 to 2023, totally 2030 images.

\paragraph{Predefined 30 text prompts for evaluation}
`A smiling man', `A smiling woman', `A man surprised', `A woman surprised', `An angry man', `An angry woman', `A man crying', `A woman crying', `A sad man', `A sad woman', `A painting of a man', `A painting of a woman', `A marble sculpture of a man', `A marble sculpture of a woman', `A 3D model of a man', `A 3D model of a woman', `A man wearing earrings', `A woman wearing earrings', `A man with a crown', `A woman with a crown', `A man wearing a cap', `A woman wearing a cap', `A man wearing eyeglasses', `A woman wearing eyeglasses', `A man with rainbow hair', `A woman with rainbow hair', `A man with shadow fade hair style', `A woman with ponytail', `A man wearing a furry cat ears headband', `A woman wearing a furry cat ears headband'.

\paragraph{Test data creation for the evaluation for the baselines (Latent Control).}
The 13 attributes that we preprocessed from FFHQ are used to measure single attribute performance. The multi-attributes that we used to measure are as followings: (`height\_between\_eyebrow\_eye', `width\_of\_face'), (`width\_of\_mouth',`height\_of\_face'), (`width\_of\_nose',`height\_of\_mouth'), (`height\_between\_mouth\_chin',`gap\_between\_eyes'), (`height\_of\_face',`height\_of\_eyes'), (`width\_of\_eyes',`width\_of\_face'), (`width\_of\_nose',`height\_of\_eyes'), (`width\_of\_nose',`gap\_between\_eyes'), (`width\_of\_eyebrow',`height\_between\_nose\_mouth'), (`height\_of\_nose',`width\_of\_eyes').
We generate a postive and a negative pair from each attribute set and compare the CR and DIS scores as mentioned in the main paper.

\paragraph{Test data creation for the evaluation for the baselines (Absolute Control).}
For each prompt, we randomly sample 50 combinations of 20 attributes. As a result, the attribute combinations for testing contains 1,500 samples; 30 (text prompts) by 50 (20-dimensional attributes combinations per prompt). Specifically, for facial composition attributes, we first obtain the means and the standard deviations of the attributes from the finetuning data. For example, the mean of `gap between eyes' is 0.687 and the standard deviation is 0.087. We sample from 2 sigma region. For age, we sample from normal distribution of which mean and standard deviation is 30 and 10. For race, we uniformly sample from 0.8 and 1 to assign the biggest value to one of 6 races. Once the major race is determined, the values for other races are randomly assigned to be sum-to-one. 

%-------------------------------------------------------------------------
\section{Additional Related Works}
\label{Appendix_sec:additional related works}
\paragraph{Controllable Text-to-Image Generation}
%Diffusion models (DMs) \cite{dhariwal2021diffusion, ho2020denoising} are state-of-art methods for image generation, which achieve remarkable high quality performance over prior generative models such as Generative Adversarial Networks (GANs) \cite{GAN} and Variational Autoencoders (VAEs) \cite{vae}. DMs operate in pixel space which make them time and resource expensive to both train and be used for inference. Latent diffusion models (LDMs)\cite{rombach2021highresolution}, e.g., Stable Diffusion, are introduced to improve on the efficiency by applying the diffusion process in the latent space while maintaining high quality image synthesis. 

T2I models \cite{rombach2021highresolution,ramesh2021zero,nichol2021glide,saharia2022photorealistic} generate high quality images from text prompts, using large pretrained visual language models like CLIP \cite{radford2021learning}. However, text only conditioning lacks fine-grained control as text often underspecifies visual details such as object type, perspective and style \cite{hutchinson2022underspecification, zhang2023iti}. %Natural languages inherently have limitations in specifying perceptual attributes (i.e., dull, colorful, skin tone) \cite{zhang2023iti}.
To improve control, various approaches incorporate images as additional conditions. ControlNet \cite{zhang2023adding} conditions image generation on inputs like depth, sketches, and semantic maps by training a copy of the diffusion model on these inputs, allowing spatial control. Similarly, T2I-Adapter \cite{mou2024t2i} uses a lightweight adapter for external signals such as images, while preserving the pre-trained model's capabilities. IP-Adapter \cite{ye2023ip-adapter} adds controllability via cross-attention networks for both text and image inputs, enabling guidance with a reference image.

However, these methods \cite{radford2021learning, zhang2023adding, mou2024t2i} generate images based on a provided example (e.g., body pose) but lack precise control over specific attributes. For instance, with a set of reference faces showing nose widths from narrowest to widest, one might want to generate an image with a specific nose width. Our approach, Att-Adapter, enables precise attribute adjustments using numeric values derived from domain-specific data (e.g., nose width ranges), allowing more accurate modifications within targeted domains.

\section{Application results: LoRA}
\label{Appendix_sec:lora application}
\begin{figure}[h]
  \vspace{-0.1in}
  \centering
    \includegraphics[width=.5\textwidth]{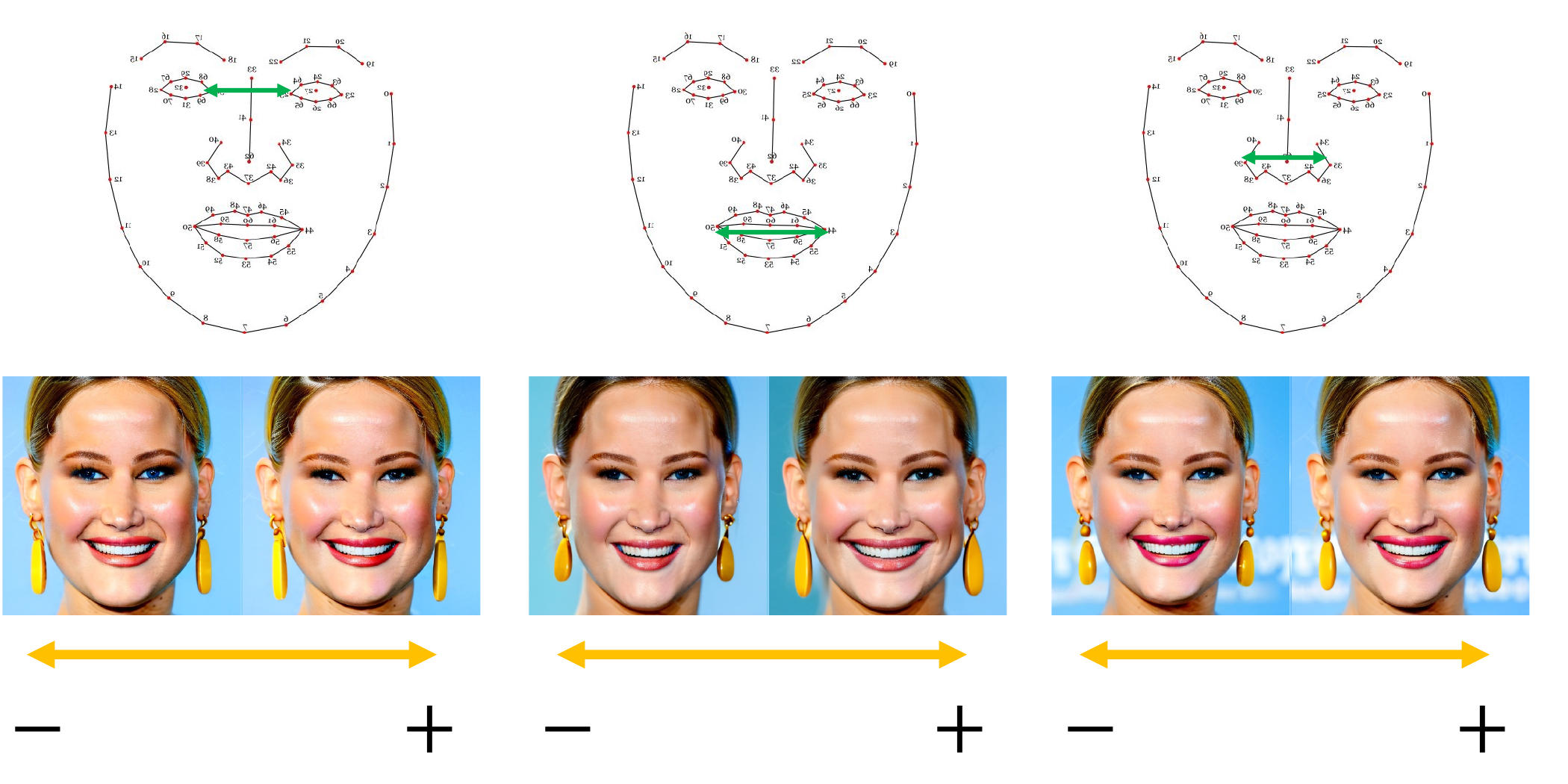}
  \caption{Qualitative experiments showing that LoRA and Att-Adapter can be combined during the sampling process.}
  \label{supfig:lora results}
    \vspace{-0.1in}
\end{figure}

LoRA can be used for personalizing pretrained Diffusion Models~\cite{hu2022lora,ruiz2023dreambooth,gal2022image}. In this section, we show that Att-Adatper can be combined with the LoRA module that is finetuned for the appearance of a specific individual. Briefly, we used 22 images of a certain celebrity (Jennifer Lawrence) to finetune LoRA. After finetuning, we combine LoRA and Att-Adapter which are separately trained. The results are shown in Fig.~\ref{supfig:lora results}. We can see that Att-Adapter can adjust the facial components of the generated image while the person identity is heavily affected by LoRA module. This shows that the wide applicability of Att-Adapter.

\section{Exploring training settings.}
\label{Appendix_sec:exploring training settings}
In this section, we explore some of the important factors that could be needed during the training process for Att-Adapter.

\paragraph{Training iteration}
First we analyze the quantitative performance per iteration which is shown in Table~\ref{table: performance per iteraton}. We can see the trade-off between the finetuned knowledge and pretrained knowledge as iteration goes on.
At 100k, we can better maintain the pretrained knowledge. However, the performance w.r.t. finetuned knowledge is not good. After 200k, even though we lose some pretrained knowledge, we get better scores for the finetuned knowledge. It is not easy to determine which checkpoint is better among 200k and 300k as their performance gap is not conspicuous. During the experiments, we empirically used the model saved at 200k.

\begin{table}[h]
\begin{adjustbox}{width=\columnwidth,center}
\begin{tabular}{cccccc}
     & \multicolumn{3}{c}{Finetuned knowledge ($\downarrow$)}              & \multicolumn{2}{c}{Pretrained knowledge ($\uparrow$)} \\ \cline{2-6} 
(Iter.)   & Facial comp.   & Age          & Race           & CLIP                   & ChatGPT         \\ \hline

100k  &     21.1           &      9.8        &       2.48         & \textbf{0.285}                       &    \textbf{82\%}             \\
% 150k  & 17.53          & 7.7         & 0.98           & 0.282                  &                 \\
200k  & \textbf{15.57}          & 8.0         & \textbf{0.36}           & 0.283                  & 78\%            \\
% 250k  & 16.29          & 7.4         & 0.46           & 0.283                  &                 \\
300k  & 15.92          & \textbf{7.7}          & \textbf{0.36}           & 0.282         &    77\%   
\end{tabular}
\end{adjustbox}
\caption{Performance per iterations. We observe that the performance is empirically converged around 200k.}
\label{table: performance per iteraton}
\end{table}

\paragraph{Embedding dimension of $Z$.}
Second, we explore the impact of dimensionality of $Z$ on performance in Table~\ref{table: performance per dimension}. Interestingly, with 128 dimension, Att-Adapter tend to learn less the finetuned knowledge while maintaining more the pretrained knowledge. With 512 dimension, the model shows decent performance in both finetuned and pretrained knowledge. With 2048 dimension, the model shows comparable performance with the 512 setting; The age prediction gets slightly better, but slightly worse in most of the other measures.

\begin{table}[h]
\begin{adjustbox}{width=\columnwidth,center}
\begin{tabular}{cccccc}
     & \multicolumn{3}{c}{Finetuned knowledge ($\downarrow$)}              & \multicolumn{2}{c}{Pretrained knowledge ($\uparrow$)} \\ \cline{2-6} 
(Dim.)   & Facial comp.   & Age          & Race           & CLIP                   & ChatGPT         \\ \hline

128  &     20.54           &      9.3        &       0.808         & \textbf{0.285}                       &    \textbf{83\%}             \\
% 150k  & 17.53          & 7.7         & 0.98           & 0.282                  &                 \\
512  & \textbf{17.42}          & 8.5         & \textbf{0.799}           & 0.282                  & 80\%            \\
% 250k  & 16.29          & 7.4         & 0.46           & 0.283                  &                 \\
2048  & 17.53          & \textbf{7.7}          & 0.987           & 0.282         &    75\%   
\end{tabular}
\end{adjustbox}
\caption{Performance comparisons given different $Z$ dimensions. For each setting, the checkpoint saved at 150k iterations is used.}
\label{table: performance per dimension}
\end{table}

\paragraph{Non-isotropic gaussian prior.}
Lastly, we compare the performance of using isotropic/non-isotropic gaussian for prior distribution. For isotropic gaussian, we estimate a scalar. For non-isotropic setting, we estimate a vector (i.e., the diagonal term of the covariance matrix.) The results are shown in Table~\ref{table: performance comparison between iso and noniso}. At 100k, we can see that the non-isotropic setting is better than the isotropic setting for learning the finetuned knowledge. We conjecture that this is because the higher degree of freedom of non-isotropic gaussian (than the isotropic gaussian) could be beneficial at fitting at some points. At 200k and 300k, however, the difference gets smaller and both settings become comparable. We empirically used the isotropic gaussian prior setting for our main experiments.

\begin{table}[h]
\begin{adjustbox}{width=\columnwidth,center}
\begin{tabular}{cccccc}
     & \multicolumn{3}{c}{Finetuned knowledge ($\downarrow$)}              & \multicolumn{2}{c}{Pretrained knowledge ($\uparrow$)} \\ \cline{2-6} 
 (Iter.)  & Facial comp.   & Age          & Race           & CLIP                   & ChatGPT         \\ \hline

100k  &     \textbf{-0.39}           &      \textbf{-0.4}        &       \textbf{-0.663}         & 0.0                       &    0\%             \\
% 150k  & 17.53          & 7.7         & 0.98           & 0.282                  &                 \\
200k  & +0.39          & \textbf{-0.2}         & +0.207           & -0.001                  & -1\%            \\
% 250k  & 16.29          & 7.4         & 0.46           & 0.283                  &                 \\
300k  & \textbf{-0.23}          & \textbf{-0.1}          & +0.076           & 0.0         &    \textbf{+2\%}   
\end{tabular}
\end{adjustbox}
\caption{Performance comparisons of two settings; 1. isotropic gaussian for the prior, 2. non-isotropic gaussian for the prior distribution. The values in the table are obtained by subtracting the values of the second setting from the values of the first setting, i.e., (informally) iso value $-$ noniso value.}
\label{table: performance comparison between iso and noniso}
\end{table}

\section{Additional comparison of Att-Adapter and LoRA}
% The better performance of Att-Adapter over the baseline in race attribute can be found in Fig.~\ref{supfig:baseline comparison on race}. 
% This can be observed by comparing the fourth and the sixth columns of (a), which shows that LoRA is confused of generating White and Hispanic.

\begin{figure}[h]
  % \vspace{-0.2in}
  \centering
    \includegraphics[width=.5\textwidth]{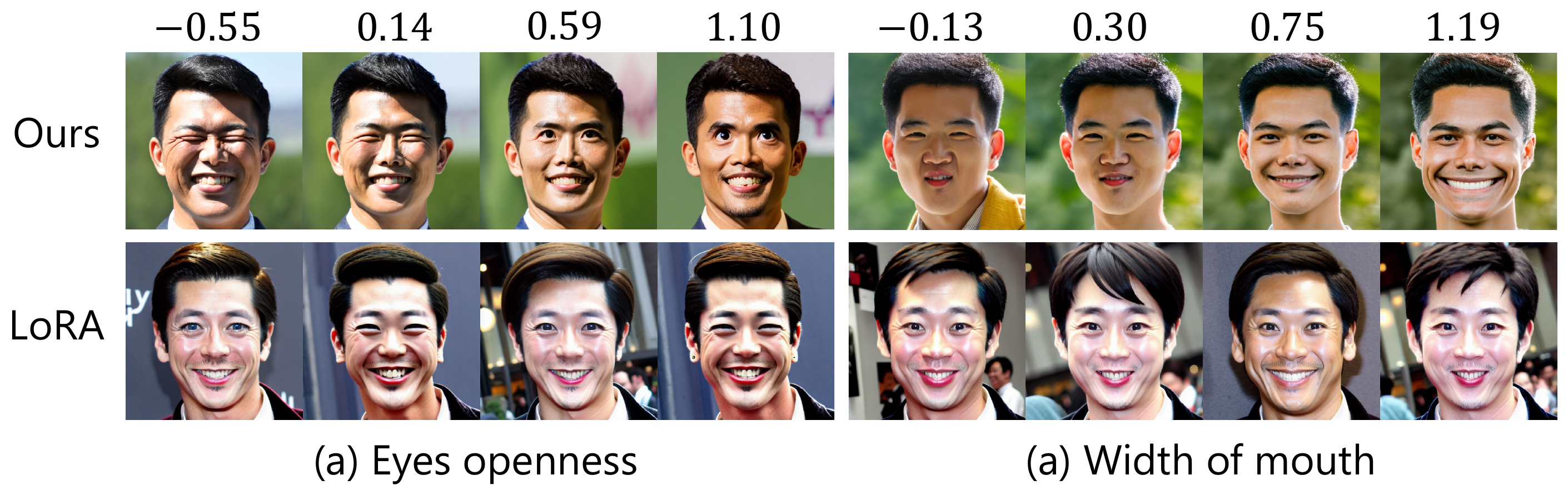}
  \caption{Extrapolation comparisons with LoRA showing the strength of Att-Adapter. A prompt of ``A stylish man smiling'' is used with `Asian' condition.}
  \label{supfig:extrapolating example}
  \vspace{-0.1in}
\end{figure}

Fig.~\ref{supfig:extrapolating example} shows the advantage of our method straightforward. Each column shows the result from the different conditioning value for the given attribute. For each macro column, we can observe from the center two rows that both Att-Adapter and LoRA show good performance given the within-domain conditioning values, i.e., $[0,1]$. However, only Att-Adapter can deal with the negative or greater-than-one conditioning. This is because LoRA is only trained with dealing with the discretized and tokenized string identifiers of 0,1,...,9. For example, given -0.55 from the leftmost column, our preprocessor for discretizing makes the value -5. We guess LoRA ignores `-' sign, and `5' is taken, which yields the eyes openness to the similar extent with the third column (i.e., 0.59). Similarly, given 1.10 in the fourth column from the left, our discretizer makes it 11, which becomes `1' and `1' after tokenized. We can see that LoRA recognizes the two `1's as `1' by comparing the eye openness with the second column (i.e., 0.14, 1 after discretized). On the other hand, as shown in the first row, Att-Adapter can extrapolate to the attribute values beyond $[0,1]$, to the unseen domain.

% \begin{figure}[h]
  
%   \centering
%     \includegraphics[width=.5\textwidth]{figures/main_comparison_race.pdf}
%   \caption{Qualitative Baseline comparisons on race. From the top, (a) LoRA, and (b) Att-Adapter. The prompts of ``A photo of a woman with smiling'' and ``A photo of a man with shadow fade hair style'' are used for the woman images and the man images.}
%   \label{supfig:baseline comparison on race}
  
% \end{figure}

\begin{figure}[h]
  % \vspace{-0.2in}
  \centering
    \includegraphics[width=.5\textwidth]{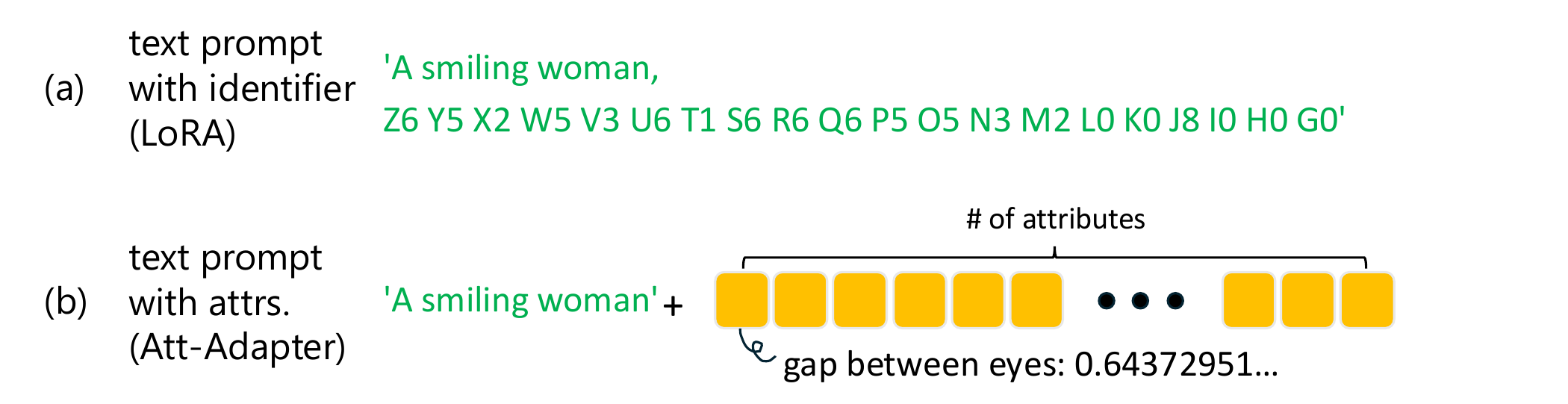}
  \caption{Visualization of the input setting of LoRA baseline. All the continuous attributes are discretized and named as special tokens. For example, the attribute `gap between eyes' and its value 0.64 becomes `Z6'. Multi-attributes are linearly converted and added consecutively.}
  \label{supfig:input settings of LoRA baseline}
  
\end{figure}

\section{Additional Analysis}
\label{Appendix_sec:additional anaylsis}

\subsection{Robustness for correlated attributes.}
\begin{figure}[t]
\vspace{-0.1in}
  \centering
  \includegraphics[width=0.45\textwidth]{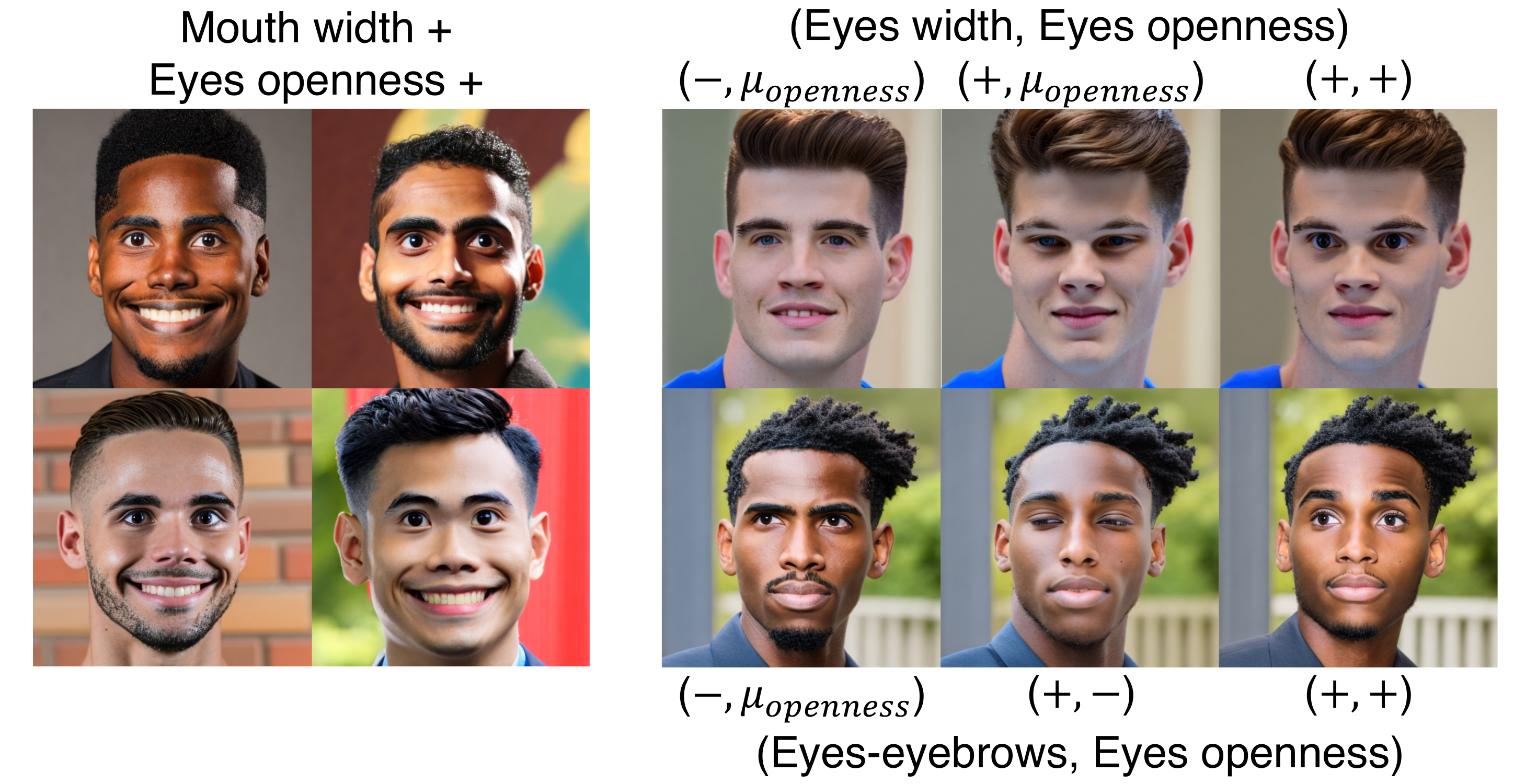}  % Replace with your image filename
  \vspace{-0.1in}
  \caption{Disentanglement of correlated attributes by Att-Adapter.}
  \vspace{-0.2in}
  %: 'eye openness' and 'mouth width' (correlated through smiling), as well 'eye width'/'eye openness' and 'eye-eyebrow distance'/'eye openness'.}
  \label{fig:disentangling}
\end{figure}

As shown in Table 2 in the main paper, Att-Adapter outperforms baselines in disentangling attributes due to joint attribute learning. %This is because Att-Adapter learns multiple attributes together, where each attribute can be learned separately, possibly mitigating the correlation issue.
%from which the correlation between attributes can be mitigated.
To provide further evidence, we performed additional analysis on correlated attributes such as "mouth width" and "eye openness", which typically co-vary through "smiling" (wider mouth coinciding with narrower eyes). As shown in Fig.~\ref{fig:disentangling} left, Att-Adapter can manipulate them independently (wider mouth + and wider eye openness +). More examples showing independent controls over the correlated attributes can be seen in Fig.~\ref{fig:disentangling} right. 

\subsection{Dependency on Quality of Attribute Annotations and Paired v.s. Unpaired Data Performance}
Att-adapter is specifically designed for scenarios where explicit paired data is unavailable but attribute annotations are present. Such scenarios frequently occur in practice, for example, product images are associated with metadata and manual annotations. %, MRI and X-ray images from different hospitals, or Cityscape images with diverse region, weather and time. %descriptive annotations in addition to manual annotations. 
 In contrast, other methods like ConceptSlider can be effective in scenarios lacking explicit attribute annotations by relying on paired data. Adapting Att-Adapter to such scenarios is nontirivial issue, which can be an interesting future direction. 

\subsection{Scalability validation}
We validated scalability by measuring resource usage with increasing number of attributes. With a batch size of 16 and latent dimension $Z\in\mathbb{R}^{1024}$, using one attribute requires around 23,704MB of VRAM and 97MB of storage. Each additional attribute increases VRAM by only about 338MB (~1.4\% of the initial VRAM) and storage by just 0.004MB (~0.004\% of the initial storage). These results support our claim of handling numerous attributes with negligible increases in memory usage.

\subsection{Ablation Study (CVAE v.s. Dropout Regularization)}
We explore the effect of dropout in preventing overfitting by conducting additional comparisons. %with dropout on ID similarity. 
%As the reviewer pointed out, dropout shows improved performance in regularizing the adapter from overfitting, which is shown as the lower ID similarity in the Table below (the best rate of 0.25 is reported). However, it still falls behind ours, verifying the better regularizing effect of CVAE.
Results in the Table below confirm using dropout (the best rate of 0.25 is reported) indeed reduces identity similarity which indicates improved regularization. However, even at this optimal dropout rate, performance remains worse than our CVAE-based approach. These confirms the effectiveness of CVAE in regularizing Att-Adapter and prevent overfitting. 

% \begin{figure}[h!]
%   \centering
%   \includegraphics[width=0.45\textwidth]{ICCV2025-Author-Kit-Feb/dropout_ablation_update.pdf}  % Replace with your image filename
%   \caption{Qualitative comparison with the dropout setting.}
%   \label{fig:dropout ablation}
% \end{figure}

\begin{table}[h!]
\centering
\vspace{-0.05in}
\begin{tabular}{ccccc}
\multicolumn{1}{l}{}  & \multicolumn{2}{c}{Naive}           & \multicolumn{1}{l}{} \\ \cline{2-3}
                      & w.o. drop & with drop & Ours                 \\ \hline
ID sim ($\downarrow$) & 28.5\%     & 23.6\%     & \textbf{6.2}\%         
\end{tabular}
\vspace{-0.05in}
\end{table}

\subsection{Challenging attributes}
\label{Appendix_subsec:challenging attributes}
\begin{figure}[h]
  \vspace{-0.1in}
  \centering
    \includegraphics[width=.5\textwidth]{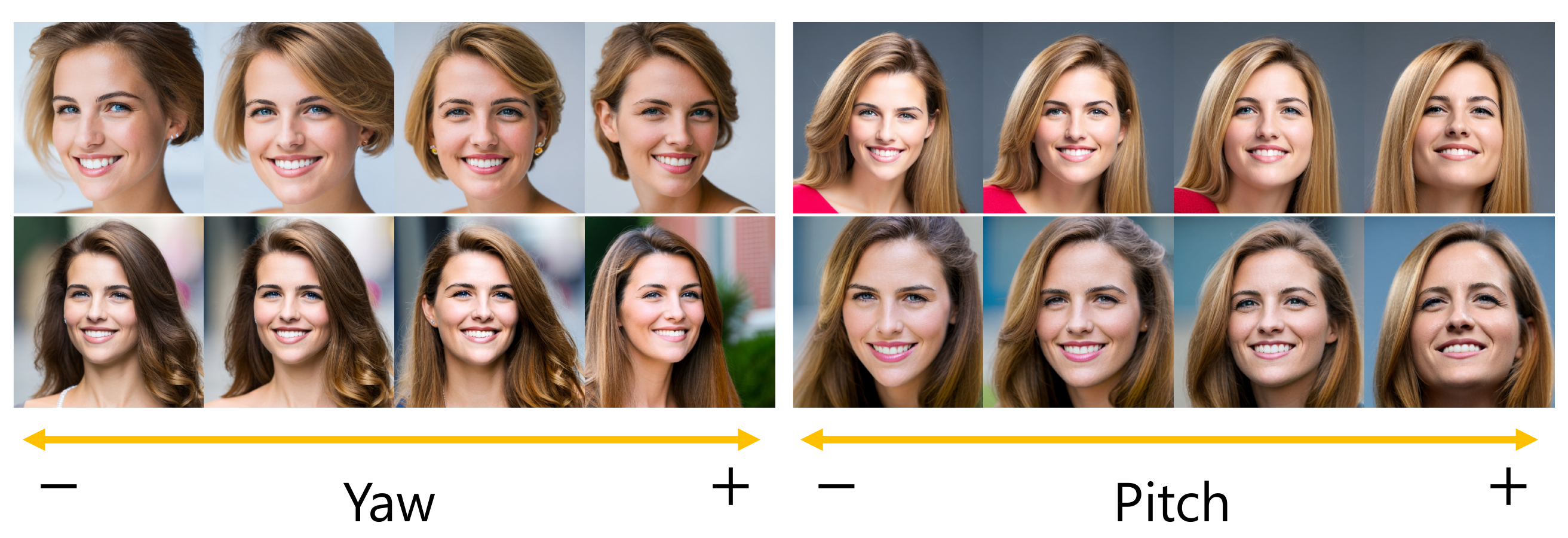}
  \caption{Qualitative results on additional attributes control. The results are generated by taking a prompt of ``A smiling woman''.}
  \label{supfig:yaw pitch results}
    \vspace{-0.1in}
\end{figure}

We show that Att-Adapter can be used beyond frontal-view images. To show this, we extend our training dataset from the frontal-view face images to the entire FFHQ dataset. We also additionally add three attributes; yaw, pitch, and roll\footnote{https://github.com/DCGM/ffhq-features-dataset}. The results are shown in Fig.~\ref{supfig:yaw pitch results}. Interestingly, even though Att-Adapter is not specifically designed for understanding 3D domain, we can see that the left-right rotation (i.e., yaw) and the up-down rotation (i.e., pitch) can be controlled. However, we observed that roll is not controllable. To improve this, we believe additional facial dataset with diverse roll information is required as FFHQ face-cropped dataset is face aligned. We also think it would be interesting and powerful if 3D domain knowledge could be aggregated in Att-Adapter which is beyond our research scope.

\begin{figure}[t]
  \vspace{-0.3in}
  \centering
    \includegraphics[width=.3\textwidth]{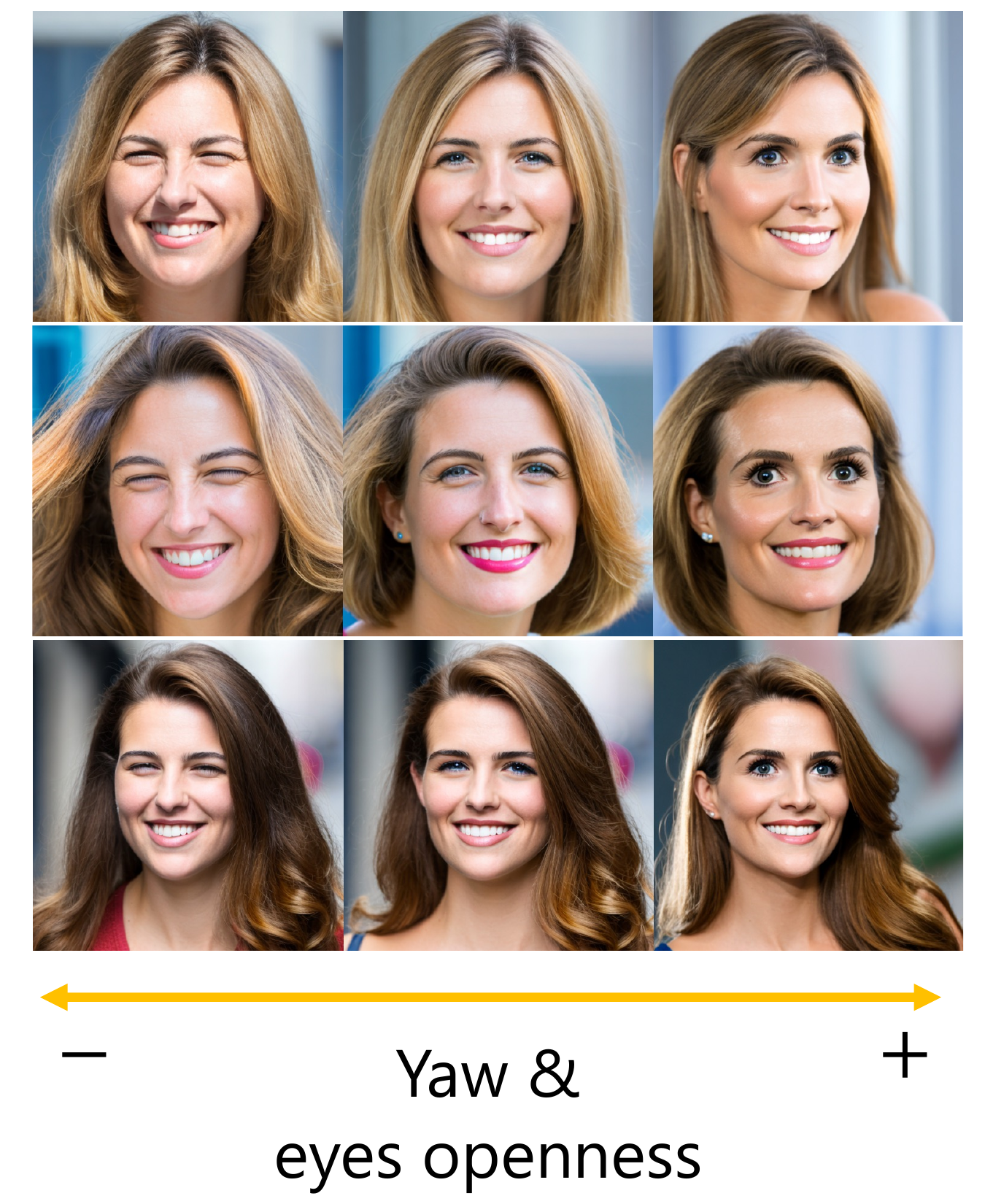}
  \caption{Additional two-attributes controlling examples. A prompt of ``A smiling woman'' is used.} 
  \label{supfig:additional multi attributes1}
    \vspace{-0.1in}
\end{figure}

Additionally, we show that Att-Adapter can control two attributes simultaneously in Fig.~\ref{supfig:additional multi attributes1}.

% \begin{figure}[h]
%   \vspace{-0.1in}
%   \centering
%     \includegraphics[width=.35\textwidth]{supp_figures/multi_attributes_supp.pdf}
%   \caption{Additional two-attributes controlling examples. A prompt of ``A smiling woman'' is used.} 
%   \label{supfig:additional multi attributes2}
%     \vspace{-0.1in}
% \end{figure}